\documentclass{article}
\usepackage{microtype}
\usepackage{graphicx}
\usepackage{subfigure}
\usepackage{booktabs} % for professional tables
\usepackage{hyperref}
\usepackage{color}

\usepackage[accepted]{icml2020}
\usepackage{amsmath,amssymb}
\usepackage{dsfont}
\usepackage{xcolor,colortbl}

\definecolor{Gray}{gray}{0.85}
\newcolumntype{a}{>{\columncolor{Gray}}c}
\definecolor{blush}{rgb}{0.87, 0.36, 0.51}
\usepackage{flushend}

\begin{document}
	\twocolumn[
	\icmltitle{Do We Really Need to Access the Source Data? Source Hypothesis Transfer for Unsupervised Domain Adaptation}
	\begin{icmlauthorlist}
		\icmlauthor{Jian Liang}{nus}
		\icmlauthor{Dapeng Hu}{nus}
		\icmlauthor{Jiashi Feng}{nus}
	\end{icmlauthorlist}
	\icmlaffiliation{nus}{Department of ECE, National University of Singapore (NUS)}
	\icmlcorrespondingauthor{Jian Liang}{liangjian92@gmail.com}
	\icmlkeywords{domain adaptation, transfer learning, hypothesis transfer, object recognition}
	\vskip 0.3in
	]
	
	\printAffiliationsAndNotice{}
	\begin{abstract}
		Unsupervised domain adaptation (UDA) aims to leverage the knowledge learned from a labeled source dataset to solve similar tasks in a new unlabeled domain.
		Prior UDA methods typically require to access the source data when learning to adapt the model, making them risky and inefficient for decentralized private data.
		This work tackles a practical setting where only a trained source model is available and investigates how we can effectively utilize such a model without source data to solve UDA problems.
		We propose a simple yet generic representation learning framework, named \emph{Source HypOthesis Transfer} (SHOT).
		SHOT freezes the classifier module (hypothesis) of the source model and learns the target-specific feature extraction module by exploiting both information maximization and self-supervised pseudo-labeling to implicitly align representations from the target domains to the source hypothesis. 
		To verify its versatility, we evaluate SHOT in a variety of adaptation cases including closed-set, partial-set, and open-set domain adaptation.
		Experiments indicate that SHOT yields state-of-the-art results among multiple domain adaptation benchmarks.
	\end{abstract}

	\section{Introduction}
	Deep neural networks have achieved remarkable success in a variety of applications across different fields but at the expense of laborious large-scale training data annotation.
	To avoid expensive data labeling, domain adaptation (DA) methods are developed to fully utilize previously labeled datasets and unlabeled data on hand in a transductive manner, which have obtained promising results in object recognition \cite{long2015learning,tzeng2017adversarial}, semantic segmentation \cite{zhang2017curriculum,hoffman2018cycada}, etc.
	Over the last decade, increasing efforts have been devoted to deep domain adaptation, especially under the vanilla unsupervised closed-set setting \cite{panareda2017open} where two domains share the same label space but the target data are not labeled. 
	One prevailing paradigm is to mitigate the distribution divergence between domains by matching the distribution statistical moments at different orders \cite{sun2016return,zellinger2017central,peng2019moment}.
	For example, the most-favored Maximum Mean Discrepancy (MMD) \cite{gretton2007kernel} measure minimizes a certain distance between weighted sums of all raw moments.
	Another popular paradigm leverages the idea of adversarial learning \cite{goodfellow2014generative} and introduces an additional domain classifier to minimize the Proxy $\mathcal{A}$\emph{-distance} \cite{ben2010theory} across domains.

	Nowadays, data are distributed on different devices and usually contain private information, e.g., those on personal phones or from surveillance cameras.
	Existing DA methods need to access the source data during learning to adapt, which is not efficient for data transmission and may violate the data privacy policy.
	In this paper, we address an \emph{interesting but challenging} unsupervised DA setting with only a trained source model provided as supervision.
	It differs from vanilla unsupervised DA in that the source model instead of the source data is provided to the unlabeled target domain.
	Such a setting helps protect the privacy in the source domain (e.g., individual hospital profiles), and transferring a light trained model is much efficient than heavy data transmission.
	For example, as shown in Table~\ref{tab:size}, the storage size of a trained model is much smaller than that of a compressed dataset.
	
	\begin{table}[!ht]
		\centering
		\small
		\vspace{-10pt}
		\caption{Storage size. The backbone networks of source models are LeNet \cite{lecun1998gradient} and ResNet-101 \cite{he2016deep}, respectively. $^\dagger$\cite{hoffman2018cycada}, $^\ddagger$\cite{peng2017visda}.}
		\label{tab:size}
		\vskip 0.05in
		%\resizebox{0.35\textwidth}{!}{$
		\begin{tabular}{lccc}
			\toprule
			Storage size (MB) & Digits$^\dagger$ & VisDA-C$^\ddagger$ \\
			\midrule
			Source Dataset & 33.2 & 7884.8 \\
			Source Model & 0.9 & 172.6 \\
			%ratio & 36.1 & 45.7 \\
			\bottomrule
		\end{tabular}
		%$}
		\vspace{0pt}
	\end{table}
	
	In terms of privacy protection, this setting seems somewhat similar to a recently proposed transfer learning setting in Hypothesis Transfer Learning (HTL) \cite{kuzborskij2013stability}, where the learner does not have direct access to the source domain data and can only operate on the hypotheses induced from the source data.
	However, our method differs significantly from that work.
	Conventional HTL always requires labeled data in the target domain \cite{tommasi2013learning} or multiple hypotheses from different source domains \cite{mansour2009domain}, which however cannot be applied to unsupervised DA.

	To address such a challenging unsupervised DA setting, we propose a simple yet generic solution called Source HypOthesis Transfer (SHOT).
	Inspired by prior work \cite{tzeng2017adversarial}, SHOT assumes that the same deep neural network model consists of a feature encoding module and a classifier module (hypothesis) across domains. 
	It aims to learn a target-specific feature encoding module to generate target data representations that are well aligned with source data representations, without accessing source data and labels for target data. 
	SHOT is designed based on the following observations. 
	If we have learned source-like representations for target data, then the classification outputs from the source classifier (hypothesis) for target data should be similar to those of source data, \emph{i.e.}, close to one-hot encodings.
	Thus, SHOT freezes the source hypothesis and fine-tunes the source encoding module by maximizing the mutual information between intermediate feature representations and outputs of the classifier, as information maximization \cite{hu2017learning} encourages the network to assign disparate one-hot encodings to the target feature representations.

	Even though information maximization forces feature representations to fit the hypothesis well, it may still align target feature representations to the wrong source hypothesis.
	To avoid this, we propose a novel self-supervised pseudo-labeling method to augment the target representation learning.
	Considering pseudo labels generated by the source classifier may be inaccurate and noisy for target data, we propose to attain intermediate class-wise prototypes for the target domain itself and further obtain cleaner pseudo labels via supervision from these prototypes to guide the mapping module learning.
	Such self-supervised labeling fully exploits the global structure of the target domain and would learn feature representations that correctly fit the source hypothesis.
	Besides, we investigate label smoothing \cite{muller2019does}, weight normalization \cite{salimans2016weight}, and batch normalization \cite{ioffe2015batch}, within the network architecture of the source model, to boost adaptation performance. Empirical evidence shows that both SHOT and its baseline method (source only) benefit from these techniques.

	We apply SHOT to the vanilla closed-set \cite{saenko2010adapting} DA scenario and a variety of other unsupervised DA tasks like the partial-set \cite{cao2018partiala} and open-set \cite{panareda2017open} cases.
	Experiments show that SHOT achieves state-of-the-art results for multiple and various domain adaptation tasks.
	For instance, on the Office-Home dataset, SHOT advances the best average accuracy from 67.6\% \cite{wang2019transferable} to \textbf{71.8\%} for the closed-set setting and from 71.8\% \cite{xu2019larger} to \textbf{79.3\%} for the partial transfer case, and from 69.5\% \cite{liu2019separate} to \textbf{72.8\%} for the open-set scenario.
	
	\section{Related Work}
	\textbf{Unsupervised Domain Adaptation.}
	Lots of unsupervised DA methods have been developed and successfully used in cross-domain applications like object recognition \cite{liang2018aggregating,liang2019exploring}, object detection \cite{chen2018domain}, semantic segmentation \cite{tsai2018learning}, and sentiment classification \cite{glorot2011domain}.
	Besides the two prevailing paradigms, moment matching and adversarial alignment, as aforementioned, there are several sample-based DA methods that estimate the importance weights of source samples \cite{jiang2007instance,huang2007correcting} or sample-to-sample similarity via optimal transport \cite{bhushan2018deepjdot}.
	Also, a few works \cite{bousmalis2016domain,ghifary2016deep} impose data reconstruction as an auxiliary task to ensure feature invariance, and some studies consider batch normalization \cite{cariucci2017autodial,wang2019transferable} and adversarial dropout \cite{saito2018adversarial,lee2019drop} within the network architecture.
	However, all these methods assume the target user's access to the source domain data, which is unsafe and sometimes unpractical since source data may be private and decentralized (stored on another device). In addition, our work is, as far as we know, the first DA framework to address almost all unsupervised DA scenarios including multi-source \cite{peng2019moment} and multi-target \cite{peng2019domain}, open-set DA \cite{panareda2017open}, and partial-set DA \cite{cao2018partiala}.
	
	\textbf{Hypothesis Transfer Learning (HTL).}
	HTL \cite{kuzborskij2013stability} can be seen as a generalization of parameter adaptation \cite{csurka2017comprehensive}, which assumes the optimal target hypothesis to be closely related to the source hypothesis.
	Like the famous fine-tuning strategy \cite{yosinski2014transferable}, HTL mostly acquires at least a small set of labeled target examples per class, limiting its applicability to the semi-supervised DA scenario \cite{yang2007cross,nelakurthi2018source}.
	Besides, a seminal work \cite{mansour2009domain} infers the weights of different source hypotheses for each unlabeled target datum and utilizes a convex combination of source hypotheses.
	However, it only fits the multi-source scenario and needs an additive distribution estimation of each source domain.
	To our best knowledge, \cite{chidlovskii2016domain,liang2019distant} introduce similar settings as ours to the unsupervised DA problem, but they may not work well enough since they do not incorporate the end-to-end feature learning module inside.

	\textbf{Pseudo Labeling (PL).}
	Pseudo labeling \cite{lee2013pseudo} is originally proposed for semi-supervised learning and gains popularity in other transductive learning problems like DA.
	The main idea is to label unlabeled data with the maximum predicted probability and perform fine-tuning together with labeled data, which is quite efficient.
	For DA methods, \cite{zhang2018collaborative,choi2019pseudo} directly incorporate pseudo labeling as a regularization while \cite{long2017deep,long2018conditional} leverage pseudo labels in the adaptation module to pursue discriminative (conditional) distribution alignment.
	\cite{zou2018unsupervised} further designs an integrated framework to alternately solve target pseudo labels and perform model training.
	In the absence of labeled data, DeepCluster \cite{caron2018deep}, one of the best self-supervised learning methods, generates pseudo labels via k-means clustering and utilizes them to re-train the current model. 
	Considering the dataset shift, our framework combines advantages of both and develops a self-supervised pseudo labeling strategy to alleviate the harm from noisy pseudo labels.
	
	\textbf{Federated Learning (FL).} FL~\cite{bonawitz2019towards} is a distributed machine learning approach that trains a model across multiple decentralized edge devices without exchanging their data samples.
	A common usage of FL consumes a simple aggregate of updates from multiple local users for the learning algorithm at the server-side
	\cite{bonawitz2017practical,mcmahan2018learning}, offering a privacy-preserving mechanism.
	Recently, \cite{peng2019federated} introduces the first federated DA setting where knowledge is transferred from the decentralized nodes to a new node without any supervision itself and proposes an adversarial-based solution.
	Specifically, it trains one model per source node and updates the target model with the aggregation of source gradients to reduce domain shift.
	However, it may fail to address the vanilla DA setting with only one source domain available.
	
	\section{Method}
	We address the unsupervised DA task with only a pre-trained source model and without access to source data. In particular, we consider $K$-way classification.
	For a vanilla unsupervised DA task, we are given $n_s$ labeled samples $\{x_s^{i},y_s^{i}\}_{i=1}^{n_s}$ from the source domain $\mathcal{D}_s$ where $x_s^{i} \in \mathcal{X}_s$, $y_s^{i}\in \mathcal{Y}_s$, and also $n_t$ unlabeled samples $\{x_t^{i}\}_{i=1}^{n_t}$ from the target domain $\mathcal{D}_t$ where $x_t^{i} \in \mathcal{X}_t$.
	The goal of DA is to predict the labels $\{y_t^{i}\}_{i=1}^{n_t}$ in the target domain, where $y_t^{i} \in \mathcal{Y}_t$, and the source task $\mathcal{X}_s \to \mathcal{Y}_s$ is assumed to be the same with the target task $\mathcal{X}_t \to \mathcal{Y}_t$.
	Here SHOT aims to learn a target function $f_t: \mathcal{X}_t \to \mathcal{Y}_t$ and infer $\{y_t^{i}\}_{i=1}^{n_t}$, with only $\{x_t^{i}\}_{i=1}^{n_t}$ and the source function $f_s: \mathcal{X}_s \to \mathcal{Y}_s$ available.

	We address the above source model transfer task for UDA through three steps. 
	First, we generate the source model from source data. Secondly, we abandon source data and transfer the model (including source hypothesis) to the target domain without accessing source data. 
	Thirdly, we further study how to design better network architectures for both models to improve adaptation performance. 
	%In the following, we elaborate on each step in detail. 
	
	\subsection{Source Model Generation}
	We consider to develop a deep neural network and learn the source model $f_s: \mathcal{X}_s \to \mathcal{Y}_s$ by minimizing the following standard cross-entropy loss,
	\begin{equation}
		\begin{aligned}
			\mathcal{L}_{src}(f_s;\mathcal{X}_s,\mathcal{Y}_s) =& \\
			-\mathbb{E}_{(x_s,y_s)\in \mathcal{X}_s \times \mathcal{Y}_s}& \sum\nolimits_{k=1}^{K} q_k \log \delta_k(f_s(x_s)),
		\end{aligned}
		\label{eq:cross}
	\end{equation}
	where $\delta_k(a)=\frac{\exp(a_k)}{\sum_i \exp(a_i)}$ denotes the $k$-th element in the soft-max output of a $K$-dimensional vector $a$, and
	$q$ is the one-of-$K$ encoding of $y_s$ where $q_k$ is `1' for the correct class and `0' for the rest.
	To further increase the discriminability of the source model and facilitate the following target data alignment, we propose to adopt the label smoothing (LS) technique as it encourages examples to lie in tight evenly separated clusters \cite{muller2019does}. 
	With LS, the objective function is changed to
	\begin{equation}
		\begin{aligned}
			\mathcal{L}_{src}^{ls}(f_s;\mathcal{X}_s,\mathcal{Y}_s) =& \\
			-\mathbb{E}_{(x_s,y_s)\in \mathcal{X}_s \times \mathcal{Y}_s}& \sum\nolimits_{k=1}^{K} q_k^{ls} \log \delta_k(f_s(x_s)),
		\end{aligned}
	\end{equation}  
	where $q^{ls}_k=(1-\alpha)q_k + \alpha/K$ is the smoothed label and $\alpha$ is the smoothing parameter which is empirically set to 0.1.
	
	\subsection{Source Hypothesis Transfer with Information Maximization (SHOT-IM)}
\begin{figure}[t]
		\centering
		\includegraphics[width=0.46\textwidth]{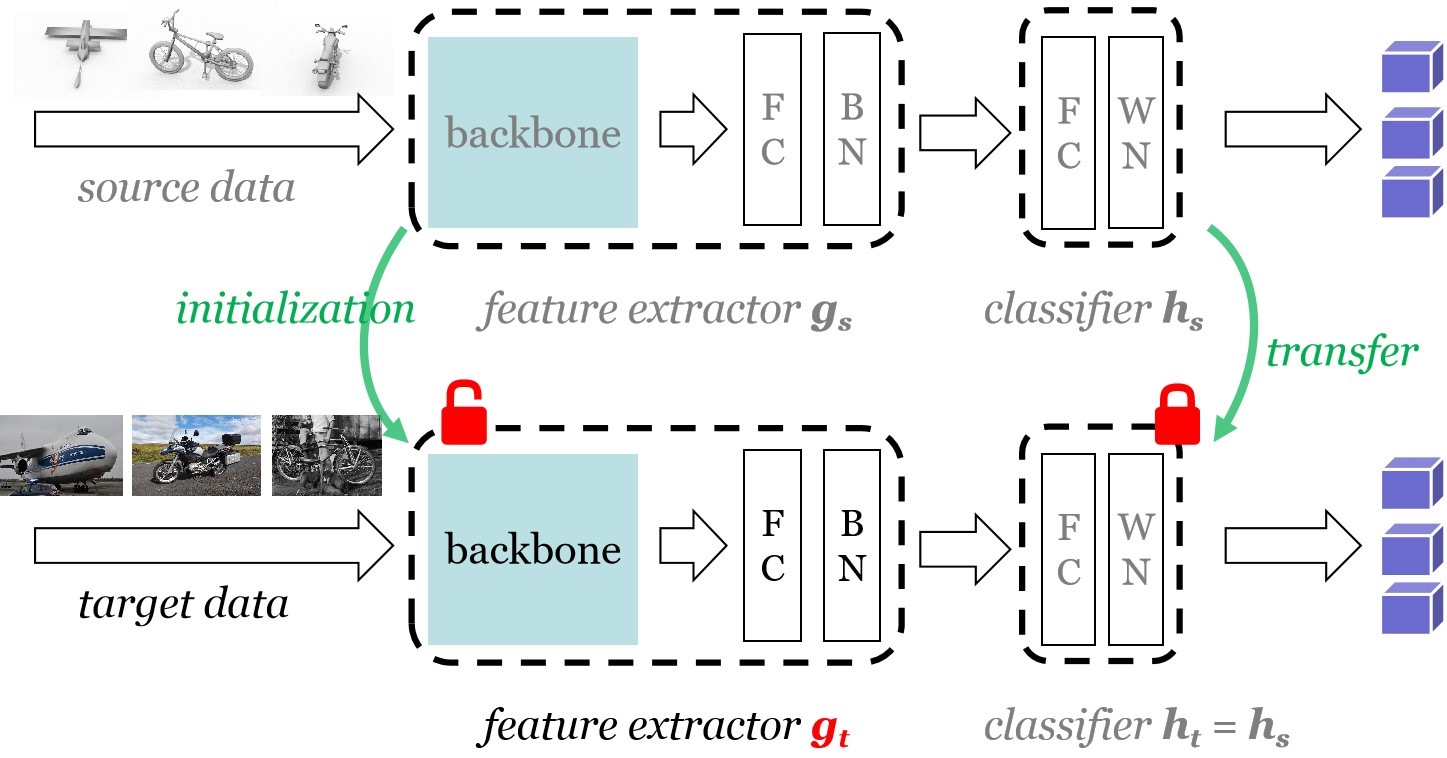}
		\caption{The pipeline of our SHOT framework. The source model consists of the feature learning module and the classifier module (hypothesis). SHOT keeps the hypothesis frozen and utilizes the feature learning module as initialization for target domain learning.}
		\label{fig:framework}
		\vspace{0pt}
	\end{figure}

	The source model parameterized by a DNN in Fig.~\ref{fig:framework} consists of two modules: the feature encoding module $g_s: \mathcal{X}_s\to \mathbb{R}^{d}$ and the classifier module $h_s: \mathbb{R}^{d}\to \mathbb{R}^{K}$, i.e., $f_s(x) = h_s\left(g_s(x)\right)$. Here $d$ is the dimension of the input feature. 
	Previous DA methods align different domains by matching the data distributions in the feature space $\mathbb{R}^{d}$ using maximum mean discrepancy (MMD) \cite{long2015learning} or domain adversarial alignment \cite{ganin2015unsupervised}.
	However, both strategies assume the source and target domains share the same feature encoder and need to access the source data during adaptation. This is not applicable in the proposed DA setting. 
	In contrast, ADDA \cite{tzeng2017adversarial} relaxes the parameter-sharing constraint and proposes a new adversarial framework for DA, which learns different mapping functions for each domain. Also, DIRT-T \cite{shu2018dirt} first trains a parameter-sharing DA framework as initialization and then fine-tunes the whole network by minimizing the cluster assumption violation.
	Both methods indicate that learning a domain-specific feature encoding module is practicable and even works better than the parameter-sharing mechanism.
	
	We follow this strategy and develop a Source HypOthesis Transfer (SHOT) framework by learning the domain-specific feature encoding module while fixing the source classifier module (hypothesis), as the source hypothesis encodes the distribution information of unseen source data.
	Specifically, SHOT uses the \emph{same} classifier module for different domain-specific feature learning modules, namely, $h_t=h_s$. It aims to learn the optimal target feature learning module $g_t: \mathcal{X}_t\to \mathbb{R}^{d}$ such that the output target features can match the source feature distribution well and can be classified accurately by the source hypothesis directly.
	It is worth noting that SHOT merely utilizes the source data for just once to generate the source hypothesis, while ADDA and DIRT-T need to access the data from source and target simultaneously when learning the model.

	Essentially, we expect to learn the optimal target encoder $g_t$ so that the target data distribution $p\left(g_t(x_t)\right)$ matches the source data distribution $p\left(g_s(x_s)\right)$ well.
	However, feature-level alignment does not work at all since it is impossible to estimate the distribution of $p\left(g_s(x_s)\right)$ without access to the source data.
	We view the challenging problem from another perspective: \emph{if the domain gap is mitigated, what kind of outputs should unlabeled target data have?}
	We argue the ideal target outputs should be similar to one-hot encoding but differ from each other.
	For this purpose, we adopt the information maximization (IM) loss \cite{krause2010discriminative,shi2012information,hu2017learning},
	to make the target outputs individually certain and globally diverse.
	In practice, we minimize the following $\mathcal{L}_{ent}$ and $\mathcal{L}_{div}$ that together constitute the IM loss: 
	\begin{equation}
		\begin{aligned}
			\mathcal{L}_{ent}(f_t;\mathcal{X}_t) &=    -\mathbb{E}_{x_t\in\mathcal{X}_t} \sum\nolimits_{k=1}^{K} \delta_k(f_t(x_t)) \log \delta_k(f_t(x_t)),\\
			\mathcal{L}_{div}(f_t;\mathcal{X}_t) &= \sum\nolimits_{k=1}^{K} \hat{p}_k \log \hat{p}_k \\
			&= D_{KL}(\hat{p}, \frac{1}{K}\mathbf{1}_K) - \log K,
		\end{aligned}
		\label{eq:ent}
	\end{equation}
	where $f_t(x)=h_t(g_t(x))$ is the $K$-dimensional output of each target sample, $\mathbf{1}_K$ is a $K$-dimensional vector with all ones, and $\hat{p} = \mathbb{E}_{x_t\in\mathcal{X}_t} [\delta(f_t(x_t))]$ is the mean output embedding of the whole target domain.
	IM would work better than conditional entropy minimization \cite{grandvalet2005semi} widely used in prior DA works \cite{vu2019advent,saito2019semi}, since IM can circumvent the trivial solution where all unlabeled data have the same one-hot encoding via the fair diversity-promoting objective $\mathcal{L}_{div}$. 
	
	\begin{figure}[t]
		\centering
		\footnotesize
		\setlength\tabcolsep{0mm}
		\renewcommand\arraystretch{0.1}
		\begin{tabular}{cc}
			\includegraphics[width=0.48\linewidth]{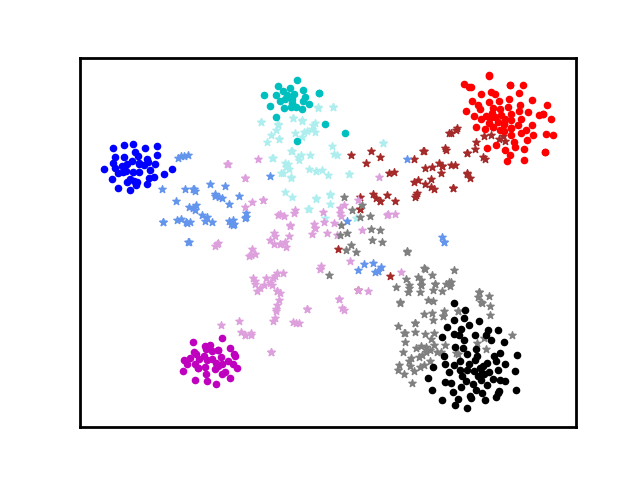} & 
			\includegraphics[width=0.48\linewidth]{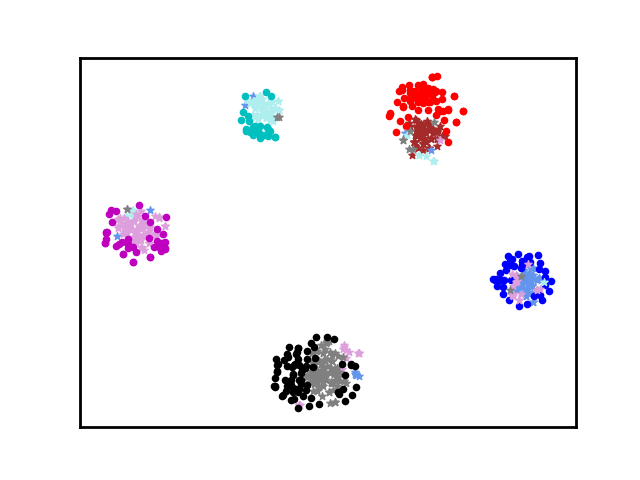} \\
			~\\
			(a) Source model only & (b) SHOT-IM 
		\end{tabular}
		\caption{t-SNE visualizations for a 5-way classification task. Circles in dark colors denote unseen source data and stars in light denote target data. Different colors represent different classes.}
		\label{fig:tsne}
	\end{figure}    
	
	\subsection{Source Hypothesis Transfer Augmented with Self-supervised Pseudo-labeling}
	\label{spl}
	As shown in Fig.~\ref{fig:tsne}, we show the t-SNE visualizations of features learned by SHOT-IM and the source model only.
	Intuitively, the target feature representations are distributed in a mess for the source model only, and using the IM loss helps align the target data with the unseen source data well.
	However, the target data may be matched to the wrong source hypothesis to some degree in Fig.~\ref{fig:tsne}(b).

	We argue that the harmful effects result from the inaccurate network outputs. For instance, a target sample from the $2$nd class with the normalized network output [0.4, 0.3, 0.1, 0.1, 0.1] may be forced to have an expected output [1.0, 0.0, 0.0, 0.0, 0.0].
	To alleviate such effects, we further apply pseudo-labeling \cite{lee2013pseudo} for each unlabeled data to better supervise the target data encoding training. 
	However, pseudo labels that are conventionally generated by source hypotheses are still noisy due to domain shift. 
	Inspired by DeepCluster \cite{caron2018deep}, we propose a novel self-supervised pseudo-labeling strategy.
	First, we attain the centroid for each class in the target domain, similar to weighted k-means clustering,
	\begin{equation}
		c_k^{(0)} = \frac{\sum_{x_t\in \mathcal{X}_t}{\delta_k(\hat{f}_t(x_t))}\ \hat{g}_t(x_t)}{\sum_{x_t\in \mathcal{X}_t}{\delta_k(\hat{f}_t(x_t))}},
		\label{eq:proto}
	\end{equation}
	where $\hat{f}_t=\hat{g}_t \circ h_t$ denotes the previously learned target hypothesis. These centroids can robustly and more reliably characterize the distribution of different categories within the target domain.
	Then, we obtain the pseudo labels via the nearest centroid classifier:
	\begin{equation}
		\hat{y}_t = \arg\min_k D_f(\hat{g}_t(x_t), c_k^{(0)}),
		\label{eq:pseudo}
	\end{equation}
	where $D_f(a,b)$ measures the cosine distance between $a$ and $b$.
	Finally, we compute the target centroids based on the new pseudo labels:
	\begin{equation}
		\begin{aligned}
			c_k^{(1)} &= \frac{\sum_{x_t\in \mathcal{X}_t}{\mathds{1}(\hat{y}_t=k)}\ \hat{g}_t(x_t)}{\sum_{x_t\in \mathcal{X}_t}{\mathds{1}(\hat{y}_t=k)}},\\
			\hat{y}_t &= \arg\min_k D_f(\hat{g}_t(x_t), c_k^{(1)}).
		\end{aligned}
		\label{eq:proto2}
	\end{equation}
	We term $\hat{y}_t$ as self-supervised pseudo labels since they are generated by the centriods obtained in an unsupervised manner.
	In practice, we update the centroids and labels in Eq.~(\ref{eq:proto2}) for multiple rounds. However, experiments verify that updating for once gives sufficiently good pseudo labels.

	To summarize, given the source model $f_s=g_s\circ h_s$ and pseudo labels generated as above, SHOT freezes the hypothesis from source $h_t=h_s$ and learns the feature encoder $g_t$ with the full objective as 
	\begin{equation}
		\begin{aligned}
			&\mathcal{L}(g_t) = \mathcal{L}_{ent} (h_t\circ g_t;\mathcal{X}_t) + \mathcal{L}_{div}(h_t\circ g_t;\mathcal{X}_t) \ - \\
			& \beta \ \mathbb{E}_{(x_t,\hat{y}_t)\in \mathcal{X}_t \times \hat{\mathcal{Y}}_t } \sum\nolimits_{k=1}^{K} \mathds{1}_{[k=\hat{y}_t]} \log \delta_k(h_t(g_t(x_t))),
		\end{aligned}
		\label{eq:overall}
	\end{equation}    
	where $\beta>0$ is a balancing hyper-parameter.

	\textbf{Remark.} The proposed SHOT framework can also be easily extended to other UDA tasks like partial-set DA \cite{cao2018partiala} and open-set DA \cite{panareda2017open}.
	More details can be found \emph{in the appendix.}
	
	\subsection{Network Architecture of Source Model}
	Now we study how to train a better source hypothesis for our own problem.
	We discuss some architecture choices for the neural network model to parameterize both the feature encoding module and hypothesis. 
	First, we need to look back at the expected network outputs for cross-entropy loss in Eq.~(\ref{eq:cross}).
	If $y_s=k$, then maximizing $f_s^{(k)}(x_s)=\frac{\exp(w_k^\top g_s(x_s))}{\sum_i \exp(w_i^\top g_s(x_s))}$ means minimizing the distance between $g_s(x_s)$ and $w_k$, where $w_k$ is the $k$-th weight vector in the last FC layer.
	Ideally, all the samples from the $k$-th class would have a feature embedding near to $w_k$. 
	If unlabeled target samples are given the correct pseudo labels, it is easily understandable that source feature embeddings are similar to target ones via the pseudo-labeling term.
	The intuition behind is quite similar to previous studies \cite{long2013transfer,xie2018learning} where a simplified MMD is exploited for discriminative domain confusion.

	Since the weight norm matters in the inner distance within the soft-max output, we adopt weight normalization (WN) \cite{salimans2016weight} to keep the norm of each weight vector $w_i$ the same in the FC classifier layer.
	Besides, as indicated in prior studies, batch normalization (BN) \cite{ioffe2015batch} can reduce the internal dataset shift since different domains share the same mean (zero) and variance which can be considered as first-order and second-order moments. 
	Based on these considerations, we form the complete framework of SHOT as shown in Fig.~\ref{fig:framework}. 
	In the experiments, we extensively study the effect of each aforementioned architecture design in Fig.~\ref{fig:acab}.
	
	\section{Experiments}
	\subsection{Setup}
	To testify its versatility, we evaluate SHOT in a variety of unsupervised DA scenarios, covering several popular visual benchmarks below.
	Full code is available at \url{https://github.com/tim-learn/SHOT}.
	
	\textbf{Office} \cite{saenko2010adapting} is a standard DA benchmark which contains three domains (Amazon (\textbf{A}), DSLR (\textbf{D}), and Webcam (\textbf{W})) and each domain consists of 31 object classes under the office environment. \cite{gong2012geodesic} further extracts 10 shared categories between \textbf{Office} and Caltech-256 (\textbf{C}), and forms a new benchmark named \textbf{Office-Caltech}.
	Both benchmarks are small-sized.

	\textbf{Office-Home} \cite{venkateswara2017deep} is a challenging medium-sized benchmark, which consists of four distinct domains (Artistic images (\textbf{Ar}), Clip Art (\textbf{Cl}), Product images (\textbf{Pr}), and Real-World images (Rw)). There are totally 65 everyday objects categories in each domain.

	\textbf{VisDA-C} \cite{peng2017visda} is a challenging large-scale benchmark that mainly focuses on the 12-class synthesis-to-real object recognition task. The source domain contains 152 thousand synthetic images generated by rendering 3D models while the target domain has 55 thousand real object images sampled from Microsoft COCO.

	\textbf{Digits} is a standard DA benchmark that focuses on digit recognition. We follow the protocol of \cite{hoffman2018cycada} and utilize three subsets: SVHN (\textbf{S}), MNIST (\textbf{M}), and USPS (\textbf{U}). Like \cite{long2018conditional}, we train our model using the training sets of each domain and report the recognition results on the standard test set of the target domain.

	\textbf{Baseline Methods.}
	For vanilla unsupervised DA in digit recognition, we compare SHOT with ADDA~\cite{tzeng2017adversarial}, ADR~\cite{saito2018adversarial}, CDAN~\cite{long2018conditional}, CyCADA~\cite{hoffman2018cycada}, CAT~\cite{deng2019cluster}, and SWD~\cite{lee2019sliced}.
	For object recognition, we compare with DANN~\cite{ganin2015unsupervised}, DAN~\cite{long2015learning}, ADR, CDAN, CAT, SAFN~\cite{xu2019larger}, BSP~\cite{chen2019transferability}, and TransNorm~\cite{wang2019transferable} and SWD.
	For specific partial-set and open-set DA tasks, we compare with IWAN~\cite{zhang2018importance}, SAN~\cite{cao2018partiala}, ETN~\cite{cao2019learning}, SAFN, and ATI-$\lambda$ \cite{panareda2017open}, OSBP~\cite{saito2018open}, OpenMax~\cite{bendale2016towards}, STA~\cite{liu2019separate}, respectively.
	For mutli-source and multi-target DA, we compare with DCTN~\cite{xu2018deep}, MCD~\cite{saito2018maximum}, M$^3$SDA-$\beta$ \cite{peng2019moment}, FADA \cite{peng2019federated}, DANN, and DADA~\cite{peng2019domain}, respectively.
	Note that results are directly cited from published papers if we follow the same setting.
	Source model only denotes using the entire mode from source for target label prediction.
	SHOT-IM is a special case of SHOT, where the self-supervised pseudo-labeling is ignored by letting $\beta=0$.

	\subsection{Implementation Details}
	\textbf{Network architecture.}
	For the digit recognition task, we use the same architectures with CDAN~\cite{long2018conditional}, namely, the classical LeNet-5 \cite{lecun1998gradient} network is utilized for USPS$\leftrightarrow$MNIST and a variant of LeNet is utilized for SVHN$\to$MNIST. Detailed networks are provided in \emph{Appendix A}.
	For the object recognition task, we employ the pre-trained ResNet-50 or ResNet-101 \cite{he2016deep} models as the backbone module like \cite{long2018conditional,deng2019cluster,xu2019larger,peng2019moment}.
	Following \cite{ganin2015unsupervised}, we replace the original FC layer with a bottleneck layer (256 units) and a task-specific FC classifier layer in Fig.~\ref{fig:framework}.
	A BN layer is put after FC inside the bottleneck layer and a weight normalization layer is utilized in the last FC layer.
	
	\setlength{\tabcolsep}{3.0pt}
	\begin{table}[tbp]
		\centering
		\small
		\caption{Classification accuracies (\%) on \textbf{Digits} dataset for \emph{vanilla closed-set DA}. S: SVHN, M:MNIST, U: USPS.}
		\label{table:digit}
		%\vskip 0.05in
		\resizebox{0.48\textwidth}{!}{$
			\begin{tabular}{lccca}
				\toprule
				Method (Source$\to$Target) & S$\to$M & U$\to$M & M$\to$U & Avg.\\
				\midrule
				Source only \cite{hoffman2018cycada} & 67.1$\pm$0.6 & 69.6$\pm$3.8 & 82.2$\pm$0.8 & 73.0 \\
				ADDA \cite{tzeng2017adversarial} & 76.0$\pm$1.8 & 90.1$\pm$0.8 & 89.4$\pm$0.2 & 85.2 \\
				ADR \cite{saito2018adversarial} & 95.0$\pm$1.9 & 93.1$\pm$1.3 & 93.2$\pm$2.5 & 93.8 \\
				CDAN+E \cite{long2018conditional} & \multicolumn{1}{l}{89.2} & \multicolumn{1}{l}{98.0} & \multicolumn{1}{l}{95.6} & 94.3 \\
				CyCADA \cite{hoffman2018cycada} & 90.4$\pm$0.4 & 96.5$\pm$0.1 & 95.6$\pm$0.4 & 94.2 \\
				rRevGrad+CAT \cite{deng2019cluster} & 98.8$\pm$0.0 & 96.0$\pm$0.9 & 94.0$\pm$0.7 & 96.3 \\
				SWD \cite{lee2019sliced} & 98.9$\pm$0.1 & 97.1$\pm$0.1 & \textbf{\color{blush}98.1}$\pm$0.1 & 98.0 \\
				\midrule
				Source model only & 70.2$\pm$1.2 & 88.0$\pm$2.2 & 79.7$\pm$2.5 & 79.3 \\
				SHOT-IM (ours) & \textbf{\color{blush}99.0}$\pm$0.1 & 97.6$\pm$0.5 & 97.7$\pm$0.1 & 98.2 \\
				SHOT (full, ours) & 98.9$\pm$0.1 & \textbf{\color{blush}98.0}$\pm$0.6 & 97.9$\pm$0.2 & \textbf{\color{blush}98.3} \\
				\midrule
				Target-supervised (Oracle) & 99.4$\pm$0.1 & 99.4$\pm$0.1 & 98.0$\pm$0.1 & 98.8 \\
				\bottomrule
			\end{tabular}
			$}
		\vspace{-5pt}
	\end{table}
	
	\textbf{Network hyper-parameters.}
	We train the whole network through back-propagation, and the newly added layers are trained with learning rate 10 times that of the pre-trained layers (backbone in Fig.~\ref{fig:framework}).
	Concretely, we adopt mini-batch SGD with momentum 0.9 and weight decay 1e$^{-3}$ and learning rate $\eta_0=1e^{-2}$ for the new layers and layers learned from scratch for all experiments except $\eta_0=1e^{-3}$ for \textbf{VisDA-C}.
	We further adopt the same learning rate scheduler $\eta=\eta_0 \cdot (1+10\cdot p)^{-0.75}$ as \cite{ganin2015unsupervised,long2018conditional}, where $p$ is the training progress changing from 0 to 1.
	Besides, we set the batch size to 64 for all the tasks. 
	We utilize $\beta=0.3$ for all experiments except $\beta=0.1$ for \textbf{Digits}.

	For \textbf{Digits}, we train the optimal source hypothesis using the test set of the source dataset as validation.
	For other datasets without train-validation splits, we randomly specify a 0.9/ 0.1 split in the source dataset and generate the optimal source hypothesis based on the validation split.
	The maximum number of epochs for Digits, Office, Office-Home, VisDA-C and Office-Caltech is empirically set as 30, 100, 50, 10 and 100, respectively.
	For learning in the target domain, we update the pseudo-labels epoch by epoch, and the maximum number of epochs is empirically set as 15.
	We randomly run our methods for three times with different random seeds $\{2019, 2020, 2021\}$ via \textbf{PyTorch} and report the average accuracies. 
	Note that we do not exploit any target augmentation such as the ten-crop ensemble \cite{long2018conditional} for evaluation.

	\subsection{Results of Digit Recognition}
	For digit recognition, we evaluate SHOT on three closed-set adaptation tasks, SVHN$\to$MNIST, USPS$\to$MNIST, and MNIST$\to$USPS.
	The classification accuracies of SHOT and prior work are reported in Table~\ref{table:digit}.
	Obviously, SHOT obtains the best or the the second-best mean accuracies for each task and outperforms prior work in terms of the average accuracy.
	Compared with source model only, SHOT-IM always achieves better results, and SHOT performs slightly better than SHOT-IM due to the contribution of self-supervised pseudo-labeling.
	It is worth noting that SHOT is comparable to the target-supervised results in M$\to$U. This may be because MNIST is much larger than USPS, which alleviates the domain shift well. 
	
	\setlength{\tabcolsep}{1.0pt}
	\begin{table}[!htbp]
		\centering
		\small
		\vspace{-10pt}
		\caption{Classification accuracies (\%) on  small-sized \textbf{Office} dataset for \emph{vanilla closed-set DA} (ResNet-50).}
		\label{table:office}
		\vskip 0.05in
		\resizebox{0.49\textwidth}{!}{$
			\begin{tabular}{lccccccacccccca}
				\toprule
				Method (Source$\to$Target) & A$\to$D & A$\to$W & D$\to$A & D$\to$W & W$\to$A & W$\to$D  & Avg. \\
				\midrule
				ResNet-50 \cite{he2016deep}     & 68.9 & 68.4 & 62.5 & 96.7 & 60.7 & 99.3 & 76.1 \\
				DANN \cite{ganin2015unsupervised} & 79.7 & 82.0  & 68.2 & 96.9 & 67.4 & 99.1 & 82.2 \\
				DAN \cite{long2015learning}      & 78.6 & 80.5 & 63.6 & 97.1 & 62.8 & 99.6 & 80.4 \\
				CDAN+E \cite{long2018conditional}  & 92.9 & 94.1 & 71.0 & 98.6 & 69.3 & \textbf{\color{blush}100.} & 87.7 \\
				rRevGrad+CAT \cite{deng2019cluster} & 90.8 & 94.4 & 72.2 & 98.0 & 70.2 & \textbf{\color{blush}100.} & 87.6 \\
				SAFN+ENT \cite{xu2019larger}    & 90.7 & 90.1 & 73.0 & 98.6 & 70.2 & 99.8 & 87.1 \\
				CDAN+BSP \cite{chen2019transferability} & 93.0 & 93.3 & 73.6 & 98.2 & 72.6 & \textbf{\color{blush}100.} & 88.5 \\
				CDAN+TransNorm \cite{wang2019transferable} & \textbf{\color{blush}94.0} & \textbf{\color{blush}95.7} & 73.4 & \textbf{\color{blush}98.7} & 74.2 & \textbf{\color{blush}100.} & \textbf{\color{blush}89.3}\\
				\midrule
				Source model only & 80.8 & 76.9 & 60.3 & 95.3 & 63.6 & 98.7 & 79.3 \\
				SHOT-IM (ours) & 90.6 & 91.2 & 72.5 & 98.3 & 71.4 & 99.9 & 87.3 \\ 
				SHOT (full, ours)   & \textbf{\color{blush}94.0} & 90.1 & \textbf{\color{blush}74.7} & 98.4 & \textbf{\color{blush}74.3} & 99.9 & 88.6 \\
				\bottomrule
			\end{tabular}
			$}
		\vspace{-5pt}
	\end{table}
	
	\setlength{\tabcolsep}{3.0pt}
	\begin{table*}[htbp]
		\centering
		\small
		%\vspace{-10pt}
		\caption{Classification accuracies (\%) on medium-sized \textbf{Office-Home} dataset for \emph{vanilla closed-set DA} (ResNet-50).}
		\label{table:home}
		\vskip 0.05in
		\resizebox{0.96\textwidth}{!}{$
			\begin{tabular}{lcccccccccccca}
				\toprule
				Method (Source$\to$Target)  & Ar$\to$Cl & Ar$\to$Pr & Ar$\to$Re & Cl$\to$Ar & Cl$\to$Pr & Cl$\to$Re & Pr$\to$Ar & Pr$\to$Cl & Pr$\to$Re & Re$\to$Ar & Re$\to$Cl & Re$\to$Pr & Avg. \\
				\midrule
				ResNet-50 \cite{he2016deep}            & 34.9 & 50.0 & 58.0 & 37.4 & 41.9 & 46.2 & 38.5 & 31.2 & 60.4 & 53.9 & 41.2 & 59.9 & 46.1 \\
				DANN \cite{ganin2015unsupervised}    & 45.6 & 59.3 & 70.1 & 47.0 & 58.5 & 60.9 & 46.1 & 43.7 & 68.5 & 63.2 & 51.8 & 76.8 & 57.6 \\
				DAN \cite{long2015learning}          & 43.6 & 57.0 & 67.9 & 45.8 & 56.5 & 60.4 & 44.0 & 43.6 & 67.7 & 63.1 & 51.5 & 74.3 & 56.3 \\
				CDAN+E \cite{long2018conditional}     & 50.7 & 70.6 & 76.0 & 57.6 & 70.0 & 70.0 & 57.4 & 50.9 & 77.3 & 70.9 & 56.7 & 81.6 & 65.8 \\
				CDAN+BSP \cite{chen2019transferability} & 52.0 & 68.6 & 76.1 & 58.0 & 70.3 & 70.2 & 58.6 & 50.2 & 77.6 & 72.2 & \textbf{\color{blush}59.3} & 81.9 & 66.3 \\
				SAFN \cite{xu2019larger}     & 52.0 & 71.7 & 76.3 & 64.2 & 69.9 & 71.9 & 63.7 & 51.4 & 77.1 & 70.9 & 57.1 & 81.5 & 67.3 \\
				CDAN+TransNorm \cite{wang2019transferable} & 50.2 & 71.4 & 77.4 & 59.3 & 72.7 & 73.1 & 61.0 & 53.1 & 79.5 & 71.9 & 59.0 & 82.9 & 67.6 \\
				\midrule
				Source model only   & 44.6 & 67.3 & 74.8 & 52.7 & 62.7 & 64.8 & 53.0 & 40.6 & 73.2 & 65.3 & 45.4 & 78.0 & 60.2 \\
				SHOT-IM (ours)   & 55.4 & 76.6 & 80.4 & 66.9 & 74.3 & 75.4 & 65.6 & 54.8 & 80.7 & \textbf{\color{blush}73.7} & 58.4 & 83.4 & 70.5 \\
				SHOT (full, ours)  & \textbf{\color{blush}57.1} & \textbf{\color{blush}78.1} & \textbf{\color{blush}81.5} & \textbf{\color{blush}68.0} & \textbf{\color{blush}78.2} & \textbf{\color{blush}78.1} & \textbf{\color{blush}67.4} & \textbf{\color{blush}54.9} & \textbf{\color{blush}82.2} & 73.3 & 58.8 & \textbf{\color{blush}84.3} & \textbf{\color{blush}71.8} \\
				\bottomrule
			\end{tabular}
			$}
		\vspace{-5pt}
	\end{table*}
	
	\setlength{\tabcolsep}{5.0pt}
	\begin{table*}[htbp]
		\centering
		\small
		\caption{Classification accuracies (\%) on  large-scale \textbf{VisDA-C} dataset for \emph{vanilla closed-set DA} (ResNet-101).}
		\label{table:visda}
		\vskip 0.05in
		\resizebox{0.9\textwidth}{!}{$
			\begin{tabular}{lcccccccccccca}
				\toprule
				Method (Synthesis $\to$ Real) & plane & bcycl & bus & car & horse & knife & mcycl & person & plant & sktbrd & train & truck & Per-class \\
				\midrule
				ResNet-101 \cite{he2016deep}  & 55.1 & 53.3 & 61.9 & 59.1 & 80.6 & 17.9 & 79.7 & 31.2  & 81.0 & 26.5  & 73.5 & 8.5  & 52.4   \\
				DANN \cite{ganin2015unsupervised}   & 81.9 & 77.7 & 82.8 & 44.3 & 81.2 & 29.5 & 65.1 & 28.6  & 51.9 & 54.6  & 82.8 & 7.8  & 57.4   \\
				DAN \cite{long2015learning}   & 87.1 & 63.0 & 76.5 & 42.0 & 90.3 & 42.9 & 85.9 & 53.1  & 49.7 & 36.3  & 85.8 & 20.7 & 61.1   \\
				ADR \cite{saito2018adversarial} & 94.2 & 48.5 & 84.0 & \textbf{\color{blush}72.9} & 90.1 & 74.2 & \textbf{\color{blush}92.6} & 72.5 & 80.8 & 61.8 & 82.2 & 28.8 & 73.5 \\
				CDAN \cite{long2018conditional}   & 85.2 & 66.9 & 83.0 & 50.8 & 84.2 & 74.9 & 88.1 & 74.5  & 83.4 & 76.0  & 81.9 & 38.0 & 73.9   \\
				CDAN+BSP \cite{chen2019transferability} & 92.4 & 61.0 & 81.0 & 57.5 & 89.0 & 80.6 & 90.1 & 77.0 & 84.2 & 77.9 & 82.1 & 38.4 & 75.9 \\
				SAFN \cite{xu2019larger} & 93.6 & 61.3 & \textbf{\color{blush}84.1} & 70.6 & \textbf{\color{blush}94.1} & 79.0 & 91.8 & 79.6 & 89.9 & 55.6 & \textbf{\color{blush}89.0} & 24.4 & 76.1 \\
				SWD \cite{lee2019sliced} & 90.8 & 82.5 & 81.7 & 70.5 & 91.7 & 69.5 & 86.3 & 77.5 & 87.4 & 63.6 & 85.6 & 29.2 & 76.4 \\
				\midrule
				Source model only & 60.9 & 21.6 & 50.9 & 67.6 & 65.8 & 6.3 & 82.2 & 23.2 & 57.3 & 30.6 & 84.6 & 8.0 & 46.6 \\
				SHOT-IM (ours) & 93.7 & 86.4 & 78.7 & 50.7 & 91.0 & 93.5 & 79.0 & 78.3 & 89.2 & 85.4 & 87.9 & 51.1 & 80.4 \\
				SHOT (full, ours)& \textbf{\color{blush}94.3} & \textbf{\color{blush}88.5} & 80.1 & 57.3 & 93.1 & \textbf{\color{blush}94.9} & 80.7 & \textbf{\color{blush}80.3} & \textbf{\color{blush}91.5} & \textbf{\color{blush}89.1} & 86.3 & \textbf{\color{blush}58.2} & \textbf{\color{blush}82.9} \\
				\bottomrule
			\end{tabular}
			$}
		\vspace{-5pt}
	\end{table*}
	
	\subsection{Results of Object Recognition (Vanilla Closed-set)}
	Next, we evaluate SHOT on a variety of object recognition benchmarks including \textbf{Office}, \textbf{Office-Home} and \textbf{VisDA-C} under the vanilla closed-set DA setting.
	As shown in Table~\ref{table:office}, SHOT performs the best for two challenging tasks, D$\to$A and W$\to$A, and performs worse than previous state-of-the-art method \cite{wang2019transferable} for other tasks. This may be because SHOT needs a relatively large target domain to learn the hypothesis $f_t$ while $D$ and $W$ are small as the target domain. Generally, SHOT obtains competitive performance even with no direct access to the source domain data.

	As expected, on the medium-sized \textbf{Office-Home} dataset, SHOT significantly outperforms previously published state-of-the-art approaches, advancing the average accuracy from 67.6\% \cite{wang2019transferable} to 71.8\% in Table~\ref{table:home}. Besides, SHOT performs the best among 10 out of 12 separate tasks.
	For a large-scale synthesis-to-real \textbf{VisDA-C} dataset in Table~\ref{table:visda}, SHOT still achieves the best per-class accuracy and performs the best among 7 out of 12 tasks.
	Carefully comparing SHOT with prior work, we find that SHOT performs well for the most challenging class `truck'.
	Again, the proposed self-supervised pseudo-labeling strategy works well as SHOT always performs better than SHOT-IM.
	
	\setlength{\tabcolsep}{3.0pt}
	\begin{table}[!htbp]
		\centering
		\small
		\vspace{-10pt}
		\caption{Average accuracies on three closed-set UDA datasets.}
		\label{tab:ab}
		\vskip 0.05in
		\resizebox{0.45\textwidth}{!}{$
			\begin{tabular}{lccc}
				\toprule
				Methods / Datasets & Office & Office-Home & VisDA-C \\
				\midrule
				Source model only & 79.3 & 60.2 & 46.6 \\
				naive pseudo-labeling (PL) \cite{lee2013pseudo} & 83.0 & 64.1 & 76.6 \\
				Self-supervised PL (ours) & 87.6 & 68.9 & 80.7\\
				\midrule
				$\mathcal{L}_{ent}$ & 83.5 & 55.5 & 63.3 \\
				$\mathcal{L}_{ent} + \mathcal{L}_{div}$ & 87.3 & 70.5 & 80.4 \\
				$\mathcal{L}_{ent} + \mathcal{L}_{div}$ + naive PL \cite{lee2013pseudo} & 87.5 & 70.3 & 82.9 \\
				$\mathcal{L}_{ent} + \mathcal{L}_{div}$ + Self-supervised PL & 88.6 & 71.8 & 82.9 \\
				%\midrule
				%$\mathcal{L}_{ent}$ + Self-supervised PL & 86.3 & 68.8 & 67.1\\
				\bottomrule
			\end{tabular}
			$}
		%\vspace{-5pt}
	\end{table}    
	
	\textbf{Ablation Study.}
	We study the advantages of the self-supervised pseudo-labeling (PL) strategy over naive one \cite{lee2013pseudo} in Table~\ref{tab:ab}.
	It is clearly shown that both PL strategies work well and our self-supervised PL is always much better than the naive one.
	Surprisingly, merely using $\mathcal{L}_{ent}$ along with self-supervised PL even produces lower results.
	But using both $\mathcal{L}_{ent}$ and $\mathcal{L}_{div}$, the results become better than the self-supervised PL baseline, which indicates 
	the importance of the diversity-promoting objective $\mathcal{L}_{div}$.
	
	\begin{figure}[t]
		\centering
		\includegraphics[width=0.3\textwidth, trim={3.5cm 7.2cm 4.2cm 8.8cm}, clip]{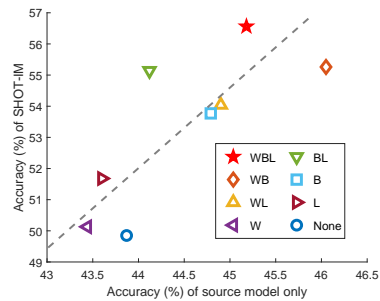}
		\vspace{-20pt}
		\caption{Accuracies (\%) on the Ar$\to$Cl task for \emph{closed-set UDA}. [\textbf{W}eight normalization/ \textbf{B}atch normalization/ \textbf{L}abel smoothing]}
		\label{fig:acab}
		\vspace{-15pt}
	\end{figure}
	
	We then evaluate the contribution of each component within the network architecture, and show the results in Fig.~\ref{fig:acab}.
	It seems that the accuracies of SHOT-IM partially depend on the quality of source model only.
	The results of source model only are especially boosted with the help of BN that is also adopted in prior studies \cite{xu2019larger,wang2019transferable}.
	Using WN and LS may decrease the accuracies of source model only, but we still obtain better results via SHOT-IM.
	All these components complement to each other, and using each two of them together helps source model only and SHOT-IM achieve the best performance.
	Finally, SHOT-IM achieves the best performance when all these components are used in the network.
	
	\setlength{\tabcolsep}{3.0pt}
	\begin{table*}[htbp]
		\centering
		\small
		\caption{(OS) Classification accuracies (\%) on \textbf{Office-Home} dataset for \emph{partial-set and open-set DA} (ResNet-50).}
		\label{table:opda}
		\vskip 0.05in
		\resizebox{0.92\textwidth}{!}{$
			\begin{tabular}{lcccccccccccca}
				\toprule
				Partial-set DA (Source$\to$Target)  & Ar$\to$Cl & Ar$\to$Pr & Ar$\to$Re & Cl$\to$Ar & Cl$\to$Pr & Cl$\to$Re & Pr$\to$Ar & Pr$\to$Cl & Pr$\to$Re & Re$\to$Ar & Re$\to$Cl & Re$\to$Pr & Avg. \\
				\midrule
				ResNet-50 \cite{he2016deep} & 46.3 & 67.5 & 75.9 & 59.1 & 59.9 & 62.7 & 58.2 & 41.8 & 74.9 & 67.4 & 48.2 & 74.2 & 61.3 \\
				IWAN \cite{zhang2018importance} & 53.9 & 54.5 & 78.1 & 61.3 & 48.0 & 63.3 & 54.2 & 52.0 & 81.3 & 76.5 & 56.8 & 82.9 & 63.6 \\
				SAN \cite{cao2018partiala}  & 44.4 & 68.7 & 74.6 & 67.5 & 65.0 & 77.8 & 59.8 & 44.7 & 80.1 & 72.2 & 50.2 & 78.7 & 65.3 \\
				ETN \cite{cao2019learning}  & 59.2 & 77.0 & 79.5 & 62.9 & 65.7 & 75.0 & 68.3 & 55.4 & 84.4 & 75.7 & 57.7 & 84.5 & 70.5 \\
				SAFN \cite{xu2019larger}  & 58.9 & 76.3 & 81.4 & 70.4 & 73.0 & 77.8 & 72.4 & 55.3 & 80.4 & 75.8 & 60.4 & 79.9 & 71.8 \\
				\midrule
				Source model only & 45.2 & 70.4 & 81.0 & 56.2 & 60.8 & 66.2 & 60.9 & 40.1 & 76.2 & 70.8 & 48.5 & 77.3 & 62.8 \\
				SHOT-IM (ours) & 57.9 & 83.6 & 88.8 & 72.4 & 74.0 & 79.0 & 76.1 & 60.6 & \textbf{\color{blush}90.1} & \textbf{\color{blush}81.9} & \textbf{\color{blush}68.3} & \textbf{\color{blush}88.5} & 76.8 \\
				SHOT (full, ours) & \textbf{\color{blush}64.8} & \textbf{\color{blush}85.2} & \textbf{\color{blush}92.7} & \textbf{\color{blush}76.3} & \textbf{\color{blush}77.6} & \textbf{\color{blush}88.8} & \textbf{\color{blush}79.7} & \textbf{\color{blush}64.3} & 89.5 & 80.6 & 66.4 & 85.8 & \textbf{\color{blush}79.3} \\
				\midrule\midrule
				Open-set DA (Source$\to$Target)  & Ar$\to$Cl & Ar$\to$Pr & Ar$\to$Re & Cl$\to$Ar & Cl$\to$Pr & Cl$\to$Re & Pr$\to$Ar & Pr$\to$Cl & Pr$\to$Re & Re$\to$Ar & Re$\to$Cl & Re$\to$Pr & Avg. \\
				\midrule
				ResNet \cite{he2016deep}  & 53.4 & 52.7 & 51.9 & 69.3 & 61.8 & 74.1 & 61.4 & 64.0 & 70.0 & 78.7 & 71.0 & 74.9 & 65.3 \\
				ATI-$\lambda$ \cite{panareda2017open} & 55.2 & 52.6 & 53.5 & 69.1 & 63.5 & 74.1 & 61.7 & 64.5 & 70.7 & 79.2 & 72.9 & 75.8 & 66.1 \\
				OSBP \cite{saito2018open}   & 56.7 & 51.5 & 49.2 & 67.5 & 65.5 & 74.0 & 62.5 & 64.8 & 69.3 & 80.6 & 74.7 & 71.5 & 65.7 \\
				OpenMax \cite{bendale2016towards}  & 56.5 & 52.9 & 53.7 & 69.1 & 64.8 & 74.5 & 64.1 & 64.0 & 71.2 & 80.3 & 73.0 & 76.9 & 66.7 \\
				STA \cite{liu2019separate}   & 58.1 & 53.1 & 54.4 & \textbf{\color{blush}71.6} & 69.3 & \textbf{\color{blush}81.9} & 63.4 & \textbf{\color{blush}65.2} & 74.9 & \textbf{\color{blush}85.0} & \textbf{\color{blush}75.8} & 80.8 & 69.5 \\
				\midrule
				Source model only & 36.3 & 54.8 & 69.1 & 33.8 & 44.4 & 49.2 & 36.8 & 29.2 & 56.8 & 51.4 & 35.1 & 62.3 & 46.6 \\
				SHOT-IM (ours) & 62.5 & 77.8 & 83.9 & 60.9 & 73.4 & 79.4 & 64.7 & 58.7 & 83.1 & 69.1 & 62.0 & 82.1 & 71.5 \\
				SHOT (full, ours) & \textbf{\color{blush}64.5} & \textbf{\color{blush}80.4} & \textbf{\color{blush}84.7} & 63.1 & \textbf{\color{blush}75.4} & 81.2 & \textbf{\color{blush}65.3} & 59.3 & \textbf{\color{blush}83.3} & 69.6 & 64.6 & \textbf{\color{blush}82.3} & \textbf{\color{blush}72.8} \\
				\midrule
			\end{tabular}
			$}
		\vspace{-10pt}
	\end{table*}    
	
	\subsection{Results of Object Recognition (Beyond Vanilla Closed-set)}
	We further extend SHOT to two other DA scenarios, multi-source~\cite{peng2019moment} and multi-target~\cite{peng2019domain}.
	For the multi-source setting, we first learn multiple source hypotheses from different source domains and directly transfer them to the target domain one by one.
	Then the prediction scores of different optimized target hypotheses via SHOT are added up to find the label which has the maximum value.
	For the multi-target setting, we naively combine these target domains and treat it as the new target domain for SHOT.
	The results of SHOT and previously published state-of-the-arts are shown in Table~\ref{table:multi}.    
	It is easy to find that SHOT achieves the best results for both settings.
	Note that FADA~\cite{peng2019federated} does not access to the source domain but utilizes the gradient update information like federated learning, but it performs much worse than our SHOT for the multi-source setting.
	
	\setlength{\tabcolsep}{3.0pt}
	\begin{table}[htbp]
		\centering
		\small
		\vspace{-10pt}
		\caption{Classification accuracies (\%) on  \textbf{Office-Caltech} dataset for \emph{multi-source and multi-target DA}~(ResNet-101). [$^*$$\mathfrak{R}$ denotes the \textbf{rest} three domains except the single source / target.]}
		\label{table:multi}
		\vskip 0.05in
		\resizebox{0.48\textwidth}{!}{$
			\begin{tabular}{lcccca}
				\toprule
				Multi-source ($\mathfrak{R}\to$) & $\mathfrak{R}\to$A & $\mathfrak{R}\to$C & $\mathfrak{R}\to$D & $\mathfrak{R}\to$W & Avg. \\
				\midrule
				ResNet-101 \cite{he2016deep} & 88.7 & 85.4 & 98.2 & 99.1 & 92.9 \\
				DAN \cite{long2015learning} & 91.6 & 89.2 & 99.1 & 99.5 & 94.8 \\
				DCTN \cite{xu2018deep}  & 92.7 & 90.2 & 99.0 & 99.4 & 95.3 \\
				MCD \cite{saito2018maximum} & 92.1 & 91.5 & 99.1 & 99.5 & 95.6 \\
				M$^3$SDA-$\beta$ \cite{peng2019moment} & 94.5 & 92.2 & \textbf{\color{blush}99.2} & 99.5 & 96.4 \\
				FADA \cite{peng2019federated}   & 84.2 & 88.7 & 87.1 & 88.1 & 87.1 \\
				\midrule
				Source model only & 95.4 & 93.7 & 98.9 & 98.3 & 96.6 \\
				SHOT-IM (ours) & 96.2 & 96.1 & 98.5 & 99.7 & 97.6 \\
				SHOT (full, ours) & \textbf{\color{blush}96.4} & \textbf{\color{blush}96.2} & 98.5 & \textbf{\color{blush}99.7} & \textbf{\color{blush}97.7} \\
				\midrule\midrule
				Multi-target ($\to \mathfrak{R}$) & A$\to\mathfrak{R}$ & C$\to\mathfrak{R}$ & D$\to\mathfrak{R}$ & W$\to\mathfrak{R}$ & Avg. \\
				\midrule
				ResNet-101 \cite{he2016deep}& 90.5 & 94.3 & 88.7 & 82.5 & 89.0 \\
				SE \cite{french2018self}   & 90.3 & 94.7 & 88.5 & 85.3 & 89.7 \\
				MCD \cite{saito2018maximum}  & 91.7 & 95.3 & 89.5 & 84.3 & 90.2 \\
				DANN \cite{ganin2015unsupervised} & 91.5 & 94.3 & 90.5 & 86.3 & 90.7 \\
				DADA \cite{peng2019domain} & 92.0 & 95.1 & 91.3 & 93.1 & 92.9 \\
				\midrule
				Source model only & 90.7 & 96.1 & 90.2 & 90.9 & 92.0 \\
				SHOT-IM (ours) & 95.7 & 97.2 & \textbf{\color{blush}96.3} & 96.1 & 96.3 \\
				SHOT (full, ours) & \textbf{\color{blush}96.2} & \textbf{\color{blush}97.3} & \textbf{\color{blush}96.3} & \textbf{\color{blush}96.2} & \textbf{\color{blush}96.5} \\
				\bottomrule
			\end{tabular}
			$}
		\vspace{-5pt}
	\end{table}
	
	Finally, we evaluate the generalization of SHOT by extending it to two challenging DA tasks, PDA and ODA, and follow the protocols on \textbf{Office-Home} in \cite{cao2019learning} and \cite{liu2019separate}, respectively. 
	For PDA, there are totally 25 classes (the first 25 in alphabetical order) in the target domain and 65 classes in the source domain, while for ODA, the source domain consists of the same 25 classes but the target domain contains 65 classes including unknown samples.
	More details about how SHOT works for these two scenarios are provided in \emph{Appendix B and C}, respectively.
	Results in Table~\ref{table:opda} validate the effectiveness of SHOT for these two challenging asymmetric DA tasks.

	\textbf{Special Case.} One may wonder whether SHOT works if we cannot train the source model by ourselves.
	To find the answer, we utilize the most popular off-the-shelf {pre-trained ImageNet models} ResNet-50 \cite{he2016deep} and consider a PDA task (\textbf{ImageNet} $\to$ \textbf{Caltech}) to evaluate the effectiveness of SHOT with the same basic setting as \cite{cao2019learning}.
	Obviously, in Table \ref{tab:ic}, SHOT still achieves slightly higher accuracy than the state-of-the-art ETN \cite{cao2019learning} even without accessing the source data.    
	
	\setlength{\tabcolsep}{3.0pt}
	\begin{table}[!htbp]
		\centering
		\vspace{-10pt}
		\caption{Results of a PDA task (\textbf{ImageNet} $\to$ \textbf{Caltech}). $^\dagger$utilizes the training set of ImageNet besides pre-trained ResNet-50 model.}
		\label{tab:ic}
		\vskip 0.05in
		\resizebox{0.45\textwidth}{!}{$
			\begin{tabular}{ccccc}
				\toprule
				Methods & ResNet-50 & ETN$^\dagger$ & SHOT-IM (ours) & SHOT (full, ours) \\
				\midrule
				Accuracy & 69.7$\pm$0.0 & 83.2$\pm$0.2 & 81.7$ \pm$0.5 & \textbf{\color{blush}83.3}$\pm$0.1\\
				\bottomrule
			\end{tabular}
			$}
		\vspace{-10pt}
	\end{table}    
	
	\section{Conclusion}
	In this paper, we address a practical unsupervised DA setting with a simple yet generic representation learning framework named SHOT.
	SHOT merely needs the well-trained source model and offers the feasibility of unsupervised DA without access to the source data which may be private and decentralized.
	Specifically, SHOT learns the optimal target-specific feature learning module to fit the source hypothesis by exploiting the information maximization and self-supervised pseudo-labeling.
	Experiments for both digit and object recognition verify that SHOT achieves competitive and even state-of-the-art performance.

	\textbf{Acknowledgements}: 
	Jiashi Feng was partially supported by AISG-100E-2019-035 and MOE2017-T2-2-151.
	The authors would like to acknowledge the (partial) support from the Open Project Program (No. KBDat1505) of Jiangsu Key Laboratory of Big Data Analysis Technology, Nanjing University of Information Science \& Technology.
	The authors also thank Quanhong Fu and Weihao Yu for their help to improve the technical writing aspect of this paper. 
	
	\begin{algorithm}[!htb]
		\caption{SHOT algorithm for closed-set UDA task.}
		\small
		\label{alg:shot}
		%\resizebox{0.46\textwidth}{!}{$
		\begin{algorithmic}
			\STATE {\bfseries Input:} source model $f_s=g_s\circ h_s$, target data $\{x_t^i\}_{i=1}^{n_t}$, maximum number of epochs $T_m$, trade-off parameter $\beta$.
			\STATE {\bfseries Initialization:} Freeze the final classifier layer $h_t = h_s$, and copy the parameters from $g_s$ to $g_t$ as initialization.
			\FOR {$epoch=1$ {\bfseries to} $T_m$}
			\STATE Obtain self-supervised pseudo labels via Eq.~(\ref{eq:proto2})
			\FOR{$iter=1$ {\bfseries to} $n_b$}
			\STATE {\color{Gray}{\# min-batch optimization}}
			\STATE Sample a batch from target data and get the corresponding pseudo labels.
			\STATE Update the parameters in $g_t$ via $\mathcal{L}(g_t)$ in Eq.~(\ref{eq:overall}).
			\ENDFOR
			\ENDFOR
		\end{algorithmic}
		%$}
	\end{algorithm}
	
	\vspace{-10pt}
	\section{Appendix}
	\vspace{0pt}
	\subsection*{A. UDA scenarios and Network Architecture}
	Concerning the DA scenarios, the closet-set setting \cite{saenko2010adapting} is the most favored, where the source and the target domain share the same label space.
	Later, the original one-to-one adaptation is extended to the multi-source \cite{peng2019moment} and the multi-target \cite{peng2019domain} setting, respectively.
	To be more realistic, open-set DA \cite{panareda2017open} assumes the target domain has some unseen classes, while partial-set DA \cite{cao2018partiala} considers a setting where the label space of the source domain subsumes that of the target domain.
	
\begin{figure}[!h]
		\centering
		\footnotesize
		\vspace{-5pt}
		\setlength\tabcolsep{1mm}
		\renewcommand\arraystretch{0.1}
		\begin{tabular}{ccc}
			\includegraphics[width=0.22\linewidth]{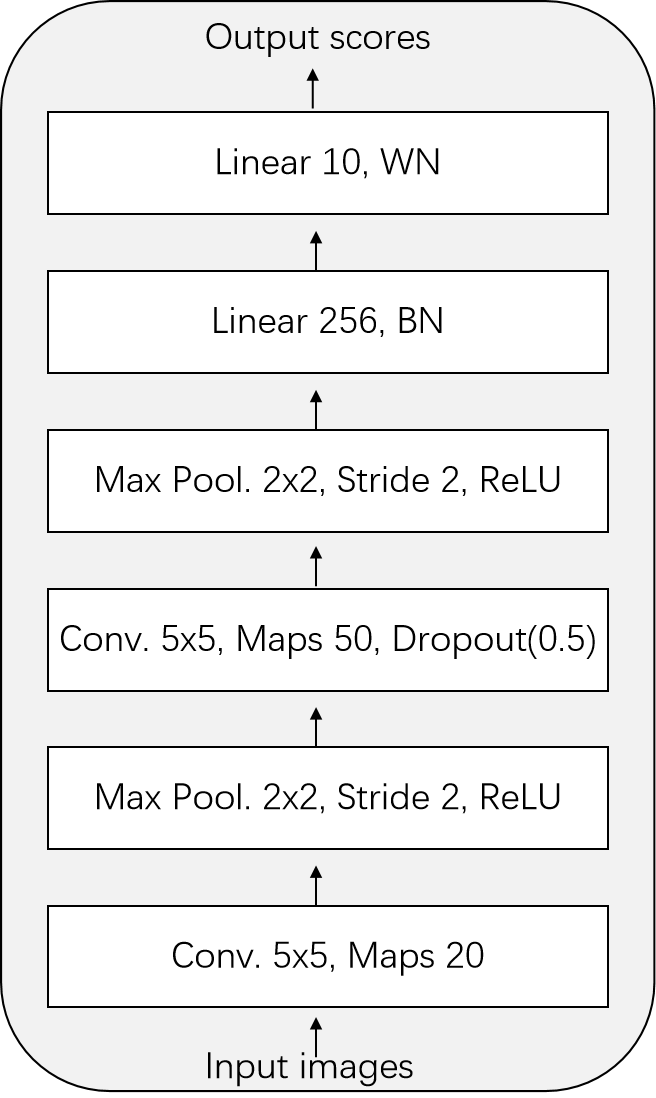} & &
			\includegraphics[width=0.26\linewidth]{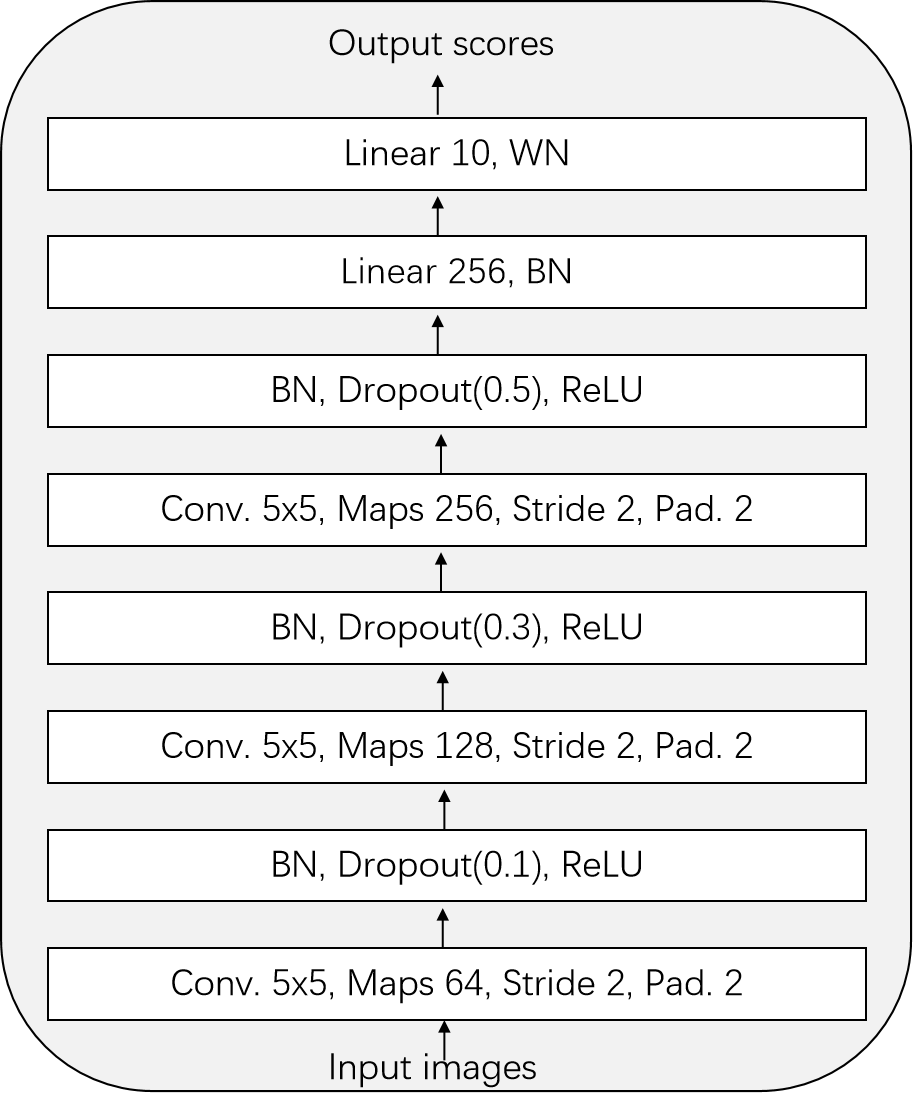} \\
			~\\
			~\\
			~\\
			(a) LeNet & & (b) DTN
		\end{tabular}
		\caption{Network architectures used for digit experiments. (a) for MNIST$\leftrightarrow$USPS and (b) for SVHN$\to$MNIST.}
		\label{fig:net}
		\vspace{-15pt}
	\end{figure}

	For the digit experiments, we follow \cite{long2018conditional} and show the detailed networks in Fig.~\ref{fig:net}.
	
	\subsection*{B. Details of SHOT for Partial-set DA (PDA)}
	Looking at the diversity-promoting term $\mathcal{L}_{div}$ in Eq.~(\ref{eq:ent}), it encourages the target domain to own a uniform target label distribution.
	It sounds reasonable for closed-set DA, but not suitable for partial-set DA.
	In reality, the target domain only contains some classes of the whole classes in the source domain, making the label distribution sparse.
	Hence, we drop the second term $\mathcal{L}_{div}$ for PDA.

\begin{figure}[!ht]
		\vspace{-10pt}
		\centering
		\includegraphics[height=1.25in, width=0.35\textwidth, trim={1.5cm 8.5cm 2.5cm 8.5cm}, clip]{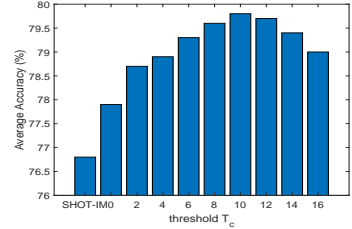}
		\vspace{-10pt}
		\caption{Average accuracies (\%) of SHOT-IM and SHOT for all partial-set DA tasks on \textbf{Office-Home}. (ResNet-50)}
		\label{fig:tc}
		\vspace{-10pt}
	\end{figure}

	Besides, within the self-supervised pseudo-labeling strategy, we usually need to obtain $K$ centroids. 
	However, for the PDA task, there are some tiny centroids which should be considered as empty like k-means clustering.
	Particularly, SHOT discards tiny centroids whose size is smaller than $T_{c}$ in Eq.~(\ref{eq:proto2}) for PDA. 
	Next, we show the average accuracy on \textbf{Office-Home} with regards to the choice of $T_c$ in Fig.~\ref{fig:tc}.
	
	\subsection*{C. Details of SHOT for Open-set DA (ODA)}
\begin{figure}[!ht]
		\centering
		\vspace{-10pt}
		\includegraphics[width=0.3\textwidth, trim={1.5cm 8.5cm 2.5cm 8.5cm}, clip]{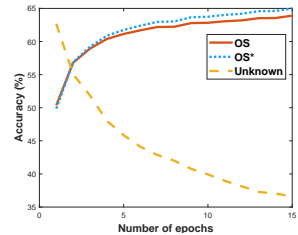}
		\vspace{-10pt}
		\caption{Different scores of SHOT w.r.t. the number of epochs for the open-set DA task Ar $\to$ Cl on \textbf{Office-Home}. (ResNet-50)}
		\label{fig:oda}
		\vspace{-10pt}
	\end{figure}

	Conversely, the target domain in an open-set DA problem contains some unknown classes besides the classes seen in the source domain.
	However, it is hard to change the final classifier layer for SHOT and learn it totally in the target domain.
	Thus we resort to a confidence thresholding strategy to reject unknown samples in the target domain during the learning of SHOT.
	In effect, we utilize the entropy of network output to compute the uncertainty, and normalize it in the range of [0,1] by dividing $\log K$.
	For each epoch, we first compute all the uncertainty values and perform a 2-class k-means clustering.
	The cluster with larger mean uncertainty would be treated as the unknown class and not be considered to update the target centroids and compute the objective in $\mathcal{L}_{ent}$. Next, we show the changes in OS score, OS* score, and unknown accuracy during the process of SHOT in Fig.~\ref{fig:oda}.
	Specifically, the OS score includes the unknown class and measures the per-class accuracy, i.e., $OS=\frac{1}{K+1}\sum_{k=1}^{K+1} Acc_k$, where $K$ indicates the number of known classes and ($K$+1)-th class is an unknown class. 
	OS* score only measure the per-class accuracy on the known classes, i.e., $OS=\frac{1}{K}\sum_{k=1}^{K} Acc_k$.
	
	\clearpage
	{
		\bibliography{example_paper}

\begin{thebibliography}{77}
\providecommand{\natexlab}[1]{#1}
\providecommand{\url}[1]{\texttt{#1}}
\expandafter\ifx\csname urlstyle\endcsname\relax
  \providecommand{\doi}[1]{doi: #1}\else
  \providecommand{\doi}{doi: \begingroup \urlstyle{rm}\Url}\fi

\bibitem[Ben-David et~al.(2010)Ben-David, Blitzer, Crammer, Kulesza, Pereira,
  and Vaughan]{ben2010theory}
Ben-David, S., Blitzer, J., Crammer, K., Kulesza, A., Pereira, F., and Vaughan,
  J.~W.
\newblock A theory of learning from different domains.
\newblock \emph{Springer MLJ}, 79\penalty0 (1-2):\penalty0 151--175, 2010.

\bibitem[Bendale \& Boult(2016)Bendale and Boult]{bendale2016towards}
Bendale, A. and Boult, T.~E.
\newblock Towards open set deep networks.
\newblock In \emph{CVPR}, 2016.

\bibitem[Bhushan~Damodaran et~al.(2018)Bhushan~Damodaran, Kellenberger,
  Flamary, Tuia, and Courty]{bhushan2018deepjdot}
Bhushan~Damodaran, B., Kellenberger, B., Flamary, R., Tuia, D., and Courty, N.
\newblock Deepjdot: Deep joint distribution optimal transport for unsupervised
  domain adaptation.
\newblock In \emph{ECCV}, 2018.

\bibitem[Bonawitz et~al.(2017)Bonawitz, Ivanov, Kreuter, Marcedone, McMahan,
  Patel, Ramage, Segal, and Seth]{bonawitz2017practical}
Bonawitz, K., Ivanov, V., Kreuter, B., Marcedone, A., McMahan, H.~B., Patel,
  S., Ramage, D., Segal, A., and Seth, K.
\newblock Practical secure aggregation for privacy-preserving machine learning.
\newblock In \emph{ACM CCS}, 2017.

\bibitem[Bonawitz et~al.(2019)Bonawitz, Eichner, Grieskamp, Huba, Ingerman,
  Ivanov, Kiddon, Konecny, Mazzocchi, McMahan, et~al.]{bonawitz2019towards}
Bonawitz, K., Eichner, H., Grieskamp, W., Huba, D., Ingerman, A., Ivanov, V.,
  Kiddon, C., Konecny, J., Mazzocchi, S., McMahan, H.~B., et~al.
\newblock Towards federated learning at scale: System design.
\newblock \emph{arXiv preprint arXiv:1902.01046}, 2019.

\bibitem[Bousmalis et~al.(2016)Bousmalis, Trigeorgis, Silberman, Krishnan, and
  Erhan]{bousmalis2016domain}
Bousmalis, K., Trigeorgis, G., Silberman, N., Krishnan, D., and Erhan, D.
\newblock Domain separation networks.
\newblock In \emph{NeurIPS}, 2016.

\bibitem[Cao et~al.(2018)Cao, Long, Wang, and Jordan]{cao2018partiala}
Cao, Z., Long, M., Wang, J., and Jordan, M.~I.
\newblock Partial transfer learning with selective adversarial networks.
\newblock In \emph{CVPR}, 2018.

\bibitem[Cao et~al.(2019)Cao, You, Long, Wang, and Yang]{cao2019learning}
Cao, Z., You, K., Long, M., Wang, J., and Yang, Q.
\newblock Learning to transfer examples for partial domain adaptation.
\newblock In \emph{CVPR}, 2019.

\bibitem[Cariucci et~al.(2017)Cariucci, Porzi, Caputo, Ricci, and
  Bulo]{cariucci2017autodial}
Cariucci, F.~M., Porzi, L., Caputo, B., Ricci, E., and Bulo, S.~R.
\newblock Autodial: Automatic domain alignment layers.
\newblock In \emph{ICCV}, 2017.

\bibitem[Caron et~al.(2018)Caron, Bojanowski, Joulin, and Douze]{caron2018deep}
Caron, M., Bojanowski, P., Joulin, A., and Douze, M.
\newblock Deep clustering for unsupervised learning of visual features.
\newblock In \emph{ECCV}, 2018.

\bibitem[Chen et~al.(2019)Chen, Wang, Long, and Wang]{chen2019transferability}
Chen, X., Wang, S., Long, M., and Wang, J.
\newblock Transferability vs. discriminability: Batch spectral penalization for
  adversarial domain adaptation.
\newblock In \emph{ICML}, 2019.

\bibitem[Chen et~al.(2018)Chen, Li, Sakaridis, Dai, and
  Van~Gool]{chen2018domain}
Chen, Y., Li, W., Sakaridis, C., Dai, D., and Van~Gool, L.
\newblock Domain adaptive faster r-cnn for object detection in the wild.
\newblock In \emph{CVPR}, 2018.

\bibitem[Chidlovskii et~al.(2016)Chidlovskii, Clinchant, and
  Csurka]{chidlovskii2016domain}
Chidlovskii, B., Clinchant, S., and Csurka, G.
\newblock Domain adaptation in the absence of source domain data.
\newblock In \emph{KDD}, 2016.

\bibitem[Choi et~al.(2019)Choi, Jeong, Kim, and Kim]{choi2019pseudo}
Choi, J., Jeong, M., Kim, T., and Kim, C.
\newblock Pseudo-labeling curriculum for unsupervised domain adaptation.
\newblock In \emph{BMVC}, 2019.

\bibitem[Csurka(2017)]{csurka2017comprehensive}
Csurka, G.
\newblock A comprehensive survey on domain adaptation for visual applications.
\newblock In \emph{Domain Adaptation in Computer Vision Applications}, pp.\
  1--35. Springer, 2017.

\bibitem[Deng et~al.(2019)Deng, Luo, and Zhu]{deng2019cluster}
Deng, Z., Luo, Y., and Zhu, J.
\newblock Cluster alignment with a teacher for unsupervised domain adaptation.
\newblock In \emph{ICCV}, 2019.

\bibitem[French et~al.(2018)French, Mackiewicz, and Fisher]{french2018self}
French, G., Mackiewicz, M., and Fisher, M.
\newblock Self-ensembling for visual domain adaptation.
\newblock In \emph{ICLR}, 2018.

\bibitem[Ganin \& Lempitsky(2015)Ganin and Lempitsky]{ganin2015unsupervised}
Ganin, Y. and Lempitsky, V.
\newblock Unsupervised domain adaptation by backpropagation.
\newblock In \emph{ICML}, 2015.

\bibitem[Ghifary et~al.(2016)Ghifary, Kleijn, Zhang, Balduzzi, and
  Li]{ghifary2016deep}
Ghifary, M., Kleijn, W.~B., Zhang, M., Balduzzi, D., and Li, W.
\newblock Deep reconstruction-classification networks for unsupervised domain
  adaptation.
\newblock In \emph{ECCV}, 2016.

\bibitem[Glorot et~al.(2011)Glorot, Bordes, and Bengio]{glorot2011domain}
Glorot, X., Bordes, A., and Bengio, Y.
\newblock Domain adaptation for large-scale sentiment classification: a deep
  learning approach.
\newblock In \emph{ICML}, 2011.

\bibitem[Gong et~al.(2012)Gong, Shi, Sha, and Grauman]{gong2012geodesic}
Gong, B., Shi, Y., Sha, F., and Grauman, K.
\newblock Geodesic flow kernel for unsupervised domain adaptation.
\newblock In \emph{CVPR}, 2012.

\bibitem[Goodfellow et~al.(2014)Goodfellow, Pouget-Abadie, Mirza, Xu,
  Warde-Farley, Ozair, Courville, and Bengio]{goodfellow2014generative}
Goodfellow, I., Pouget-Abadie, J., Mirza, M., Xu, B., Warde-Farley, D., Ozair,
  S., Courville, A., and Bengio, Y.
\newblock Generative adversarial nets.
\newblock In \emph{NeurIPS}, 2014.

\bibitem[Grandvalet \& Bengio(2005)Grandvalet and Bengio]{grandvalet2005semi}
Grandvalet, Y. and Bengio, Y.
\newblock Semi-supervised learning by entropy minimization.
\newblock In \emph{NeurIPS}, 2005.

\bibitem[Gretton et~al.(2007)Gretton, Borgwardt, Rasch, Sch{\"o}lkopf, and
  Smola]{gretton2007kernel}
Gretton, A., Borgwardt, K., Rasch, M., Sch{\"o}lkopf, B., and Smola, A.~J.
\newblock A kernel method for the two-sample-problem.
\newblock In \emph{NeurIPS}, 2007.

\bibitem[He et~al.(2016)He, Zhang, Ren, and Sun]{he2016deep}
He, K., Zhang, X., Ren, S., and Sun, J.
\newblock Deep residual learning for image recognition.
\newblock In \emph{CVPR}, 2016.

\bibitem[Hoffman et~al.(2018)Hoffman, Tzeng, Park, Zhu, Isola, Saenko, Efros,
  and Darrell]{hoffman2018cycada}
Hoffman, J., Tzeng, E., Park, T., Zhu, J.-Y., Isola, P., Saenko, K., Efros, A.,
  and Darrell, T.
\newblock Cycada: Cycle-consistent adversarial domain adaptation.
\newblock In \emph{ICML}, 2018.

\bibitem[Hu et~al.(2017)Hu, Miyato, Tokui, Matsumoto, and
  Sugiyama]{hu2017learning}
Hu, W., Miyato, T., Tokui, S., Matsumoto, E., and Sugiyama, M.
\newblock Learning discrete representations via information maximizing
  self-augmented training.
\newblock In \emph{ICML}, 2017.

\bibitem[Huang et~al.(2007)Huang, Gretton, Borgwardt, Sch{\"o}lkopf, and
  Smola]{huang2007correcting}
Huang, J., Gretton, A., Borgwardt, K., Sch{\"o}lkopf, B., and Smola, A.~J.
\newblock Correcting sample selection bias by unlabeled data.
\newblock In \emph{NeurIPS}, 2007.

\bibitem[Ioffe \& Szegedy(2015)Ioffe and Szegedy]{ioffe2015batch}
Ioffe, S. and Szegedy, C.
\newblock Batch normalization: Accelerating deep network training by reducing
  internal covariate shift.
\newblock In \emph{ICML}, 2015.

\bibitem[Jiang \& Zhai(2007)Jiang and Zhai]{jiang2007instance}
Jiang, J. and Zhai, C.
\newblock Instance weighting for domain adaptation in nlp.
\newblock In \emph{ACL}, 2007.

\bibitem[Krause et~al.(2010)Krause, Perona, and
  Gomes]{krause2010discriminative}
Krause, A., Perona, P., and Gomes, R.~G.
\newblock Discriminative clustering by regularized information maximization.
\newblock In \emph{NeurIPS}, 2010.

\bibitem[Kuzborskij \& Orabona(2013)Kuzborskij and
  Orabona]{kuzborskij2013stability}
Kuzborskij, I. and Orabona, F.
\newblock Stability and hypothesis transfer learning.
\newblock In \emph{ICML}, 2013.

\bibitem[LeCun et~al.(1998)LeCun, Bottou, Bengio, and
  Haffner]{lecun1998gradient}
LeCun, Y., Bottou, L., Bengio, Y., and Haffner, P.
\newblock Gradient-based learning applied to document recognition.
\newblock \emph{Proceedings of the IEEE}, 86\penalty0 (11):\penalty0
  2278--2324, 1998.

\bibitem[Lee et~al.(2019{\natexlab{a}})Lee, Batra, Baig, and
  Ulbricht]{lee2019sliced}
Lee, C.-Y., Batra, T., Baig, M.~H., and Ulbricht, D.
\newblock Sliced wasserstein discrepancy for unsupervised domain adaptation.
\newblock In \emph{CVPR}, 2019{\natexlab{a}}.

\bibitem[Lee(2013)]{lee2013pseudo}
Lee, D.-H.
\newblock Pseudo-label: The simple and efficient semi-supervised learning
  method for deep neural networks.
\newblock In \emph{Workshop on challenges in representation learning, ICML},
  2013.

\bibitem[Lee et~al.(2019{\natexlab{b}})Lee, Kim, Kim, and Jeong]{lee2019drop}
Lee, S., Kim, D., Kim, N., and Jeong, S.-G.
\newblock Drop to adapt: Learning discriminative features for unsupervised
  domain adaptation.
\newblock In \emph{ICCV}, 2019{\natexlab{b}}.

\bibitem[Liang et~al.(2018)Liang, He, Sun, and Tan]{liang2018aggregating}
Liang, J., He, R., Sun, Z., and Tan, T.
\newblock Aggregating randomized clustering-promoting invariant projections for
  domain adaptation.
\newblock \emph{IEEE TPAMI}, 41\penalty0 (5):\penalty0 1027--1042, 2018.

\bibitem[Liang et~al.(2019)Liang, He, Sun, and Tan]{liang2019distant}
Liang, J., He, R., Sun, Z., and Tan, T.
\newblock Distant supervised centroid shift: A simple and efficient approach to
  visual domain adaptation.
\newblock In \emph{CVPR}, 2019.

\bibitem[Liu et~al.(2019)Liu, Cao, Long, Wang, and Yang]{liu2019separate}
Liu, H., Cao, Z., Long, M., Wang, J., and Yang, Q.
\newblock Separate to adapt: Open set domain adaptation via progressive
  separation.
\newblock In \emph{CVPR}, 2019.

\bibitem[Long et~al.(2013)Long, Wang, Ding, Sun, and Yu]{long2013transfer}
Long, M., Wang, J., Ding, G., Sun, J., and Yu, P.~S.
\newblock Transfer feature learning with joint distribution adaptation.
\newblock In \emph{ICCV}, 2013.

\bibitem[Long et~al.(2015)Long, Cao, Wang, and Jordan]{long2015learning}
Long, M., Cao, Y., Wang, J., and Jordan, M.
\newblock Learning transferable features with deep adaptation networks.
\newblock In \emph{ICML}, 2015.

\bibitem[Long et~al.(2017)Long, Zhu, Wang, and Jordan]{long2017deep}
Long, M., Zhu, H., Wang, J., and Jordan, M.~I.
\newblock Deep transfer learning with joint adaptation networks.
\newblock In \emph{ICML}, 2017.

\bibitem[Long et~al.(2018)Long, Cao, Wang, and Jordan]{long2018conditional}
Long, M., Cao, Z., Wang, J., and Jordan, M.~I.
\newblock Conditional adversarial domain adaptation.
\newblock In \emph{NeurIPS}, 2018.

\bibitem[Mansour et~al.(2009)Mansour, Mohri, and
  Rostamizadeh]{mansour2009domain}
Mansour, Y., Mohri, M., and Rostamizadeh, A.
\newblock Domain adaptation with multiple sources.
\newblock In \emph{NeurIPS}, 2009.

\bibitem[McMahan et~al.(2018)McMahan, Ramage, Talwar, and
  Zhang]{mcmahan2018learning}
McMahan, H.~B., Ramage, D., Talwar, K., and Zhang, L.
\newblock Learning differentially private recurrent language models.
\newblock In \emph{ICLR}, 2018.

\bibitem[M{\"u}ller et~al.(2019)M{\"u}ller, Kornblith, and
  Hinton]{muller2019does}
M{\"u}ller, R., Kornblith, S., and Hinton, G.~E.
\newblock When does label smoothing help?
\newblock In \emph{NeurIPS}, 2019.

\bibitem[Nelakurthi et~al.(2018)Nelakurthi, Maciejewski, and
  He]{nelakurthi2018source}
Nelakurthi, A.~R., Maciejewski, R., and He, J.
\newblock Source free domain adaptation using an off-the-shelf classifier.
\newblock In \emph{IEEE BigData}, 2018.

\bibitem[Panareda~Busto \& Gall(2017)Panareda~Busto and Gall]{panareda2017open}
Panareda~Busto, P. and Gall, J.
\newblock Open set domain adaptation.
\newblock In \emph{ICCV}, 2017.

\bibitem[Peng et~al.(2017)Peng, Usman, Kaushik, Hoffman, Wang, and
  Saenko]{peng2017visda}
Peng, X., Usman, B., Kaushik, N., Hoffman, J., Wang, D., and Saenko, K.
\newblock Visda: The visual domain adaptation challenge.
\newblock \emph{arXiv preprint arXiv:1710.06924}, 2017.

\bibitem[Peng et~al.(2019{\natexlab{a}})Peng, Bai, Xia, Huang, Saenko, and
  Wang]{peng2019moment}
Peng, X., Bai, Q., Xia, X., Huang, Z., Saenko, K., and Wang, B.
\newblock Moment matching for multi-source domain adaptation.
\newblock In \emph{ICCV}, 2019{\natexlab{a}}.

\bibitem[Peng et~al.(2019{\natexlab{b}})Peng, Huang, Sun, and
  Saenko]{peng2019domain}
Peng, X., Huang, Z., Sun, X., and Saenko, K.
\newblock Domain agnostic learning with disentangled representations.
\newblock In \emph{ICML}, 2019{\natexlab{b}}.

\bibitem[Peng et~al.(2020)Peng, Huang, Zhu, and Saenko]{peng2019federated}
Peng, X., Huang, Z., Zhu, Y., and Saenko, K.
\newblock Federated adversarial domain adaptation.
\newblock In \emph{ICLR}, 2020.

\bibitem[Saenko et~al.(2010)Saenko, Kulis, Fritz, and
  Darrell]{saenko2010adapting}
Saenko, K., Kulis, B., Fritz, M., and Darrell, T.
\newblock Adapting visual category models to new domains.
\newblock In \emph{ECCV}, 2010.

\bibitem[Saito et~al.(2018{\natexlab{a}})Saito, Ushiku, Harada, and
  Saenko]{saito2018adversarial}
Saito, K., Ushiku, Y., Harada, T., and Saenko, K.
\newblock Adversarial dropout regularization.
\newblock In \emph{ICLR}, 2018{\natexlab{a}}.

\bibitem[Saito et~al.(2018{\natexlab{b}})Saito, Watanabe, Ushiku, and
  Harada]{saito2018maximum}
Saito, K., Watanabe, K., Ushiku, Y., and Harada, T.
\newblock Maximum classifier discrepancy for unsupervised domain adaptation.
\newblock In \emph{CVPR}, 2018{\natexlab{b}}.

\bibitem[Saito et~al.(2018{\natexlab{c}})Saito, Yamamoto, Ushiku, and
  Harada]{saito2018open}
Saito, K., Yamamoto, S., Ushiku, Y., and Harada, T.
\newblock Open set domain adaptation by backpropagation.
\newblock In \emph{ECCV}, 2018{\natexlab{c}}.

\bibitem[Saito et~al.(2019)Saito, Kim, Sclaroff, Darrell, and
  Saenko]{saito2019semi}
Saito, K., Kim, D., Sclaroff, S., Darrell, T., and Saenko, K.
\newblock Semi-supervised domain adaptation via minimax entropy.
\newblock In \emph{ICCV}, 2019.

\bibitem[Salimans \& Kingma(2016)Salimans and Kingma]{salimans2016weight}
Salimans, T. and Kingma, D.~P.
\newblock Weight normalization: A simple reparameterization to accelerate
  training of deep neural networks.
\newblock In \emph{NeurIPS}, 2016.

\bibitem[Shi \& Sha(2012)Shi and Sha]{shi2012information}
Shi, Y. and Sha, F.
\newblock Information-theoretical learning of discriminative clusters for
  unsupervised domain adaptation.
\newblock In \emph{ICML}, 2012.

\bibitem[Shu et~al.(2018)Shu, Bui, Narui, and Ermon]{shu2018dirt}
Shu, R., Bui, H.~H., Narui, H., and Ermon, S.
\newblock A dirt-t approach to unsupervised domain adaptation.
\newblock In \emph{ICLR}, 2018.

\bibitem[Sun et~al.(2016)Sun, Feng, and Saenko]{sun2016return}
Sun, B., Feng, J., and Saenko, K.
\newblock Return of frustratingly easy domain adaptation.
\newblock In \emph{AAAI}, 2016.

\bibitem[Tommasi et~al.(2013)Tommasi, Orabona, and Caputo]{tommasi2013learning}
Tommasi, T., Orabona, F., and Caputo, B.
\newblock Learning categories from few examples with multi model knowledge
  transfer.
\newblock \emph{IEEE TPAMI}, 36\penalty0 (5):\penalty0 928--941, 2013.

\bibitem[Tsai et~al.(2018)Tsai, Hung, Schulter, Sohn, Yang, and
  Chandraker]{tsai2018learning}
Tsai, Y.-H., Hung, W.-C., Schulter, S., Sohn, K., Yang, M.-H., and Chandraker,
  M.
\newblock Learning to adapt structured output space for semantic segmentation.
\newblock In \emph{CVPR}, 2018.

\bibitem[Tzeng et~al.(2017)Tzeng, Hoffman, Saenko, and
  Darrell]{tzeng2017adversarial}
Tzeng, E., Hoffman, J., Saenko, K., and Darrell, T.
\newblock Adversarial discriminative domain adaptation.
\newblock In \emph{CVPR}, 2017.

\bibitem[Venkateswara et~al.(2017)Venkateswara, Eusebio, Chakraborty, and
  Panchanathan]{venkateswara2017deep}
Venkateswara, H., Eusebio, J., Chakraborty, S., and Panchanathan, S.
\newblock Deep hashing network for unsupervised domain adaptation.
\newblock In \emph{CVPR}, 2017.

\bibitem[Vu et~al.(2019)Vu, Jain, Bucher, Cord, and P{\'e}rez]{vu2019advent}
Vu, T.-H., Jain, H., Bucher, M., Cord, M., and P{\'e}rez, P.
\newblock Advent: Adversarial entropy minimization for domain adaptation in
  semantic segmentation.
\newblock In \emph{CVPR}, 2019.

\bibitem[Wang et~al.(2019)Wang, Jin, Long, Wang, and
  Jordan]{wang2019transferable}
Wang, X., Jin, Y., Long, M., Wang, J., and Jordan, M.~I.
\newblock Transferable normalization: Towards improving transferability of deep
  neural networks.
\newblock In \emph{NeurIPS}, 2019.

\bibitem[Xie et~al.(2018)Xie, Zheng, Chen, and Chen]{xie2018learning}
Xie, S., Zheng, Z., Chen, L., and Chen, C.
\newblock Learning semantic representations for unsupervised domain adaptation.
\newblock In \emph{ICML}, 2018.

\bibitem[Xu et~al.(2018)Xu, Chen, Zuo, Yan, and Lin]{xu2018deep}
Xu, R., Chen, Z., Zuo, W., Yan, J., and Lin, L.
\newblock Deep cocktail network: Multi-source unsupervised domain adaptation
  with category shift.
\newblock In \emph{CVPR}, 2018.

\bibitem[Xu et~al.(2019)Xu, Li, Yang, and Lin]{xu2019larger}
Xu, R., Li, G., Yang, J., and Lin, L.
\newblock Larger norm more transferable: An adaptive feature norm approach for
  unsupervised domain adaptation.
\newblock In \emph{ICCV}, 2019.

\bibitem[Yang et~al.(2007)Yang, Yan, and Hauptmann]{yang2007cross}
Yang, J., Yan, R., and Hauptmann, A.~G.
\newblock Cross-domain video concept detection using adaptive svms.
\newblock In \emph{ACM-MM}, 2007.

\bibitem[Yosinski et~al.(2014)Yosinski, Clune, Bengio, and
  Lipson]{yosinski2014transferable}
Yosinski, J., Clune, J., Bengio, Y., and Lipson, H.
\newblock How transferable are features in deep neural networks?
\newblock In \emph{NeurIPS}, 2014.

\bibitem[Zellinger et~al.(2017)Zellinger, Grubinger, Lughofer, Natschl{\"a}ger,
  and Saminger-Platz]{zellinger2017central}
Zellinger, W., Grubinger, T., Lughofer, E., Natschl{\"a}ger, T., and
  Saminger-Platz, S.
\newblock Central moment discrepancy (cmd) for domain-invariant representation
  learning.
\newblock In \emph{ICLR}, 2017.

\bibitem[Zhang et~al.(2018{\natexlab{a}})Zhang, Ding, Li, and
  Ogunbona]{zhang2018importance}
Zhang, J., Ding, Z., Li, W., and Ogunbona, P.
\newblock Importance weighted adversarial nets for partial domain adaptation.
\newblock In \emph{CVPR}, 2018{\natexlab{a}}.

\bibitem[Zhang et~al.(2018{\natexlab{b}})Zhang, Ouyang, Li, and
  Xu]{zhang2018collaborative}
Zhang, W., Ouyang, W., Li, W., and Xu, D.
\newblock Collaborative and adversarial network for unsupervised domain
  adaptation.
\newblock In \emph{CVPR}, 2018{\natexlab{b}}.

\bibitem[Zhang et~al.(2017)Zhang, David, and Gong]{zhang2017curriculum}
Zhang, Y., David, P., and Gong, B.
\newblock Curriculum domain adaptation for semantic segmentation of urban
  scenes.
\newblock In \emph{ICCV}, 2017.

\bibitem[Zou et~al.(2018)Zou, Yu, Vijaya~Kumar, and Wang]{zou2018unsupervised}
Zou, Y., Yu, Z., Vijaya~Kumar, B., and Wang, J.
\newblock Unsupervised domain adaptation for semantic segmentation via
  class-balanced self-training.
\newblock In \emph{ECCV}, 2018.

\end{thebibliography}
		\bibliographystyle{icml2020}
	}
	
\end{document}